%% file: emnlp.tex
\documentclass[11pt,a4paper]{article}
\usepackage[hyperref]{emnlp2020}
\usepackage{times}
\usepackage{latexsym}

\usepackage{microtype}

\aclfinalcopy 
\def\aclpaperid{3126} 

\usepackage{bm}
\usepackage{amsthm}
\usepackage{array}
\usepackage{algorithm}
\usepackage{algpseudocode}
\usepackage{amsmath}
\usepackage{amssymb}
\usepackage{booktabs}
\usepackage{bm}
\usepackage{color}
\usepackage{bbm}
\usepackage{etex}
\usepackage{graphicx}
\usepackage{multirow}
\usepackage{stmaryrd}
\usepackage{subcaption}
\usepackage{url}
\usepackage{thmtools}
\usepackage{mathtools}
\usepackage{ctable}

\newcommand{\squishlist}{
	\begin{list}{$\bullet$}
		{ \setlength{\itemsep}{0pt}
			\setlength{\parsep}{3pt}
			\setlength{\topsep}{3pt}
			\setlength{\partopsep}{0pt}
			\setlength{\leftmargin}{1.5em}
			\setlength{\labelwidth}{1em}
			\setlength{\labelsep}{0.5em} } }
	
	\newcommand{\squishlisttwo}{
		\begin{list}{$\bullet$}
			{ \setlength{\itemsep}{0pt}
				\setlength{\parsep}{0pt}
				\setlength{\topsep}{0pt}
				\setlength{\partopsep}{0pt}
				\setlength{\leftmargin}{2em}
				\setlength{\labelwidth}{1.5em}
				\setlength{\labelsep}{0.5em} } }
		
		\newcommand{\squishend}{
	\end{list}  }

\newcommand{\specialcell}[2][c]{%
  \begin{tabular}[#1]{@{}c@{}}#2\end{tabular}}

\title{{HABERTOR}: {A}n {E}fficient and {E}ffective {D}eep {H}atespeech {De}tector}

\author{Thanh Tran\texttt{$^\bigstar$}, 
	Yifan Hu\texttt{$^\diamondsuit$}, 
	Changwei Hu\texttt{$^\diamondsuit$}, 
	Kevin Yen\texttt{$^\diamondsuit$}, 
	Fei Tan\texttt{$^\diamondsuit$}, 
	Kyumin Lee\texttt{$^\bigstar$}, 
	Serim Park\texttt{$^\heartsuit$}\\
	\texttt{$^\bigstar$}Worcester Polytechnic Institute, 
	\texttt{$^\diamondsuit$}Yahoo! Research,
	\texttt{$^\heartsuit$}Twitter \\
	\texttt{\{tdtran,kmlee\}@wpi.edu} \\
	\texttt{\{yifanhu,changweih,kevinyen,fei.tan\}@verizonmedia.com} \\
	\texttt{serimp@twitter.com }  \\
}

\date{}

\begin{document}
\maketitle
\begin{abstract}
We present our {HABERTOR} model for detecting hatespeech in large scale user-generated content. Inspired by the recent success of the BERT model, we propose several modifications to BERT to enhance the performance on the downstream hatespeech classification task. {HABERTOR} inherits BERT's architecture, but is different in four aspects: (i) it generates its own vocabularies and is pre-trained from the scratch using the largest scale hatespeech dataset; (ii) it consists of \emph{Quaternion-based factorized} components, resulting in a much smaller number of parameters, faster training and inferencing, as well as less memory usage; (iii) it uses our proposed multi-source ensemble heads with a pooling layer for separate input sources, to further enhance its effectiveness; and (iv) it uses a regularized adversarial training with our proposed fine-grained and adaptive noise magnitude to enhance its robustness.
Through experiments on the large-scale real-world hatespeech dataset with 1.4M annotated comments, we show that {HABERTOR} works better than \textbf{15} state-of-the-art hatespeech detection methods, including fine-tuning Language Models. In particular, comparing with BERT, our {HABERTOR} is 4$\sim$5 times faster in the training/inferencing phase, uses less than 1/3 of the memory, and has better performance, even though we pre-train it by using less than 1\% of the number of words. Our generalizability analysis shows that {HABERTOR} transfers well to other unseen hatespeech datasets and is a more efficient and effective alternative to BERT for the hatespeech classification. 
\end{abstract}

\input{1-intro.tex}

\input{2-relatedwork.tex}

\input{3-problem.tex}

\input{4-approach.tex}

\input{5-experiments.tex}
\input{7-Conclusion.tex}

\bibliographystyle{acl_natbib}
\bibliography{emnlp}

\clearpage
\appendix
\input{8-supplemental}

\end{document}

%% file: 1-intro.tex
\section{Introduction}





The occurrence of hatespeech has been increasing~\cite{hate_speech_increasing}. It has become easier than before to reach a large audience quickly via social media, causing an increase of the temptation for inappropriate behaviors such as hatespeech, and potential damage to social systems. In particular, hatespeech interferes with civil discourse and turns good people away. Furthermore, hatespeech in the virtual world can lead to physical violence against certain groups in the real world\footnote{https://www.nytimes.com/2018/10/31/opinion/caravan-hate-speech-bowers-sayoc.html}\footnote{https://www.washingtonpost.com/nation/2018/11/30/how-online-hate-speech-is-fueling-real-life-violence}, so it should not be ignored on the ground of freedom of speech.

To detect hatespeech, researchers developed human-crafted feature-based classifiers \cite{chatzakou2017mean,davidson2017automated,waseem2016hateful,macavaney2019hate}, and proposed deep neural network architectures \cite{zampieri2019semeval,gamback2017using,park2017one,badjatiya2017deep,agrawal2018deep}. 
However, they might not explore all possible important features for hatespeech detection, ignored pre-trained language model understanding, or proposed uni-directional language models by reading from left to right or right to left.


Recently, the BERT (Bidirectional Encoder Representations from Transformers) model \cite{devlin2018bert} has achieved tremendous success in Natural Language Processing 
. The key innovation of BERT is in applying the transformer~\cite{vaswani2017attention} to language modeling tasks. 
A BERT model pre-trained on these language modeling tasks forms a good basis for further fine-tuning on supervised tasks such as machine translation and question answering, \emph{etc}.

Recent work on hatespeech detection \cite{nikolov2019nikolov} has applied the BERT model and has shown its prominent results over previous hatespeech classifiers. However, we point out its two limitations in hatespeech detection domain. First, the previous studies \cite{elsherief2018peer,elsherief2018hate} have shown that a hateful corpus owns distinguished linguistic/semantic characteristics compared to a non-hateful corpus. For instance, hatespeech sequences are often informal or even
intentionally mis-spelled~\cite{elsherief2018hate,arango2019hate}, so words in hateful sequences can sit in a long tail when ranking their uniqueness, and a comment can be hateful or non-hateful using the same words \cite{zhang2019hate}. 
For example, ``dick'' in the sentence ``Nobody knew dick about what that meant'' is non-hateful, but ``d1ck'' in ``You are a weak small-d1cked keyboard warrior'' is hateful \footnote{\textbf{It is important to note that this paper contains hate speech examples, which may be offensive to some readers. They do not represent the views of the authors. We tried to make a balance between showing less number of hate speech examples and illustrating the challenges in real-world applications.}}. 
Thus, to better understand hateful vocabularies and contexts, it is better to pre-train on a mixture of both hateful and non-hateful corpora. Doing so helps to overcome the
limitation of using BERT models pre-trained on non-hateful corpora like English Wikipedia and BookCorpus. Second, even the smallest pre-trained BERT ``base'' model contains 110M parameters. It takes a lot of computational resources to pre-train, fine-tune, and serve. 
Some recent efforts aim to reduce 
the complexity of BERT model with the knowledge distillation technique such as DistillBert \cite{sanh2019distilbert} and TinyBert \cite{jiao2019tinybert}. In these methods, a pre-trained BERT-alike model is used as a teacher model, and a student (smaller) model (i.e. TinyBERT, DistilBERT, \emph{.etc}) is trained to produce similar output to that of the teacher model. Unfortunately, while their complexity is reduced, the performance is also degraded in NLP tasks compared to BERT. Another direction is to use cross-layer parameter sharing, such as ALBERT \cite{Lan2020ALBERT}. However, ALBERT's computational time is similar to BERT, since the number of layers remains the same as BERT; likewise, its inference is equally expensive.

Based on the above observation and analysis, we aim to investigate whether it is possible to achieve a better hatespeech prediction performance than state-of-the-art machine learning classifiers, including classifiers based on publicly available BERT model, while significantly reducing the number of parameters compared with the BERT model. By doing so, we believe that performing pre-training tasks from the ground up and on a hatespeech-related corpus would allow the model to understand hatespeech patterns better and enhance the predictive results. However, while language model pretraining tasks require a large scale corpus size, available hatespeech datasets are normally small: only 16K$\sim$115K annotated comments \cite{waseem2016hateful,wulczyn2017ex}. Thus, we introduce a large annotated hatespeech dataset with 1.4M comments extracted from Yahoo News and Yahoo Finance. To reduce the complexity, we reduce the number of layers and hidden size, and propose Quaternion-based Factorization mechanisms in BERT architecture. To further improve the model effectiveness and robustness, we introduce a multi-source ensemble-head fine-tuning architecture, as well as a target-based adversarial training.

The major contributions of our work are:
\squishlist

\item We reduce the number of parameters in BERT considerably, and consequently the training/inferencing time and memory, while achieving better performance compared to the much larger BERT models, and other state-of-the-art hatespeech detection methods.



\item We pre-train from the ground up a hateful language model with our proposed Quaternion Factorization methods on a large-scale hatespeech dataset, which gives better performance than fine tuning a pretrained BERT model. 
\item We propose a flexible classification net with multi-sources and multi-heads, building on top of the learned sequence representations to further enhance our model's predictive capability.
\item We utilize adversarial training with a proposed fine-grained and adaptive noise magnitude to improve our model's performance.
\squishend


%% file: 2-relatedwork.tex
\section{Related Work}
\label{sec:related}

Some of the earlier works in hatespeech detection have applied a variety of classical machine learning algorithms \cite{chatzakou2017mean,davidson2017automated,waseem2016hateful,macavaney2019hate}. Their intuition is to do feature engineering (i.e. manually generate features), then apply classification methods such as SVM, Random Forest, and Logistic Regression. The features are mostly Term-Frequency Inverse-Document-Frequency scores or Bag-of-Words vectors, and can be combined with additional features extracted from the user account's meta information and network structure (i.e., followers, followees, \emph{etc}). Those methods are suboptimal as they mainly rely on the quality and quantity of the human-crafted features.

Recent works have used deep neural network architectures for hatespeech detection \cite{zampieri2019semeval,mouswe2} such as CNN \cite{gamback2017using,park2017one}, RNN (i.e. LSTM and GRU) \cite{badjatiya2017deep,agrawal2018deep}, combining CNN with RNN \cite{zhang2018detecting}, or fine tuning a pretrained language models \cite{indurthi-etal-2019-fermi}.


Another direction 
focuses on the testing generalization of the current hatespeech classifiers \cite{agrawal2018deep,dadvar2018cyberbullying,grondahl2018all}, where those methods are tested in other datasets and domains such as Twitter data \cite{waseem2016hateful}, Wikipedia data \cite{wulczyn2017ex}, Formspring data \cite{reynolds2011using}, and YouTube comment data \cite{dadvar2014experts}.

Unlike previous works, we pre-train a hateful language model, then build a multi-source multi-head hatespeech classifier with regularized adversarial training to enhance the model's performance.



%% file: 3-problem.tex
\vspace{-5pt}
\section{Problem Definition}
\label{sec:problem}
Given an input text sequence $s = [w_1, w_2, ..., w_n]$ where \{$w_1$, $w_2$, .., $w_n$\} are words and $n = |s|$ is the maximum length of the input sequence $s$. 
The hatespeech classification task aims to build a mapping function $f: s=[w_1, w_2, ..., w_n] \longrightarrow \mathcal{R} \in [0, 1]$, where $f$ inputs $s$ and returns a probability score $P(y=1 | s) \in [0, 1]$, indicating how likely $s$ is classified as hatespeech. In this paper, we approximate $f$ by a deep neural classifier, where we first pretrain $f$ with unsupervised language modeling tasks to enhance its language understanding. Then, we train $f$ with the hatespeech classification task to produce $P(y=1 | s)$. 


%% file: 4-approach.tex
\section{Our approach -- HABERTOR}
\label{sec:approach}

\begin{figure*}[t]
    \centering
	\begin{subfigure}{0.41\textwidth}
		\centering
		\includegraphics[width=\textwidth]{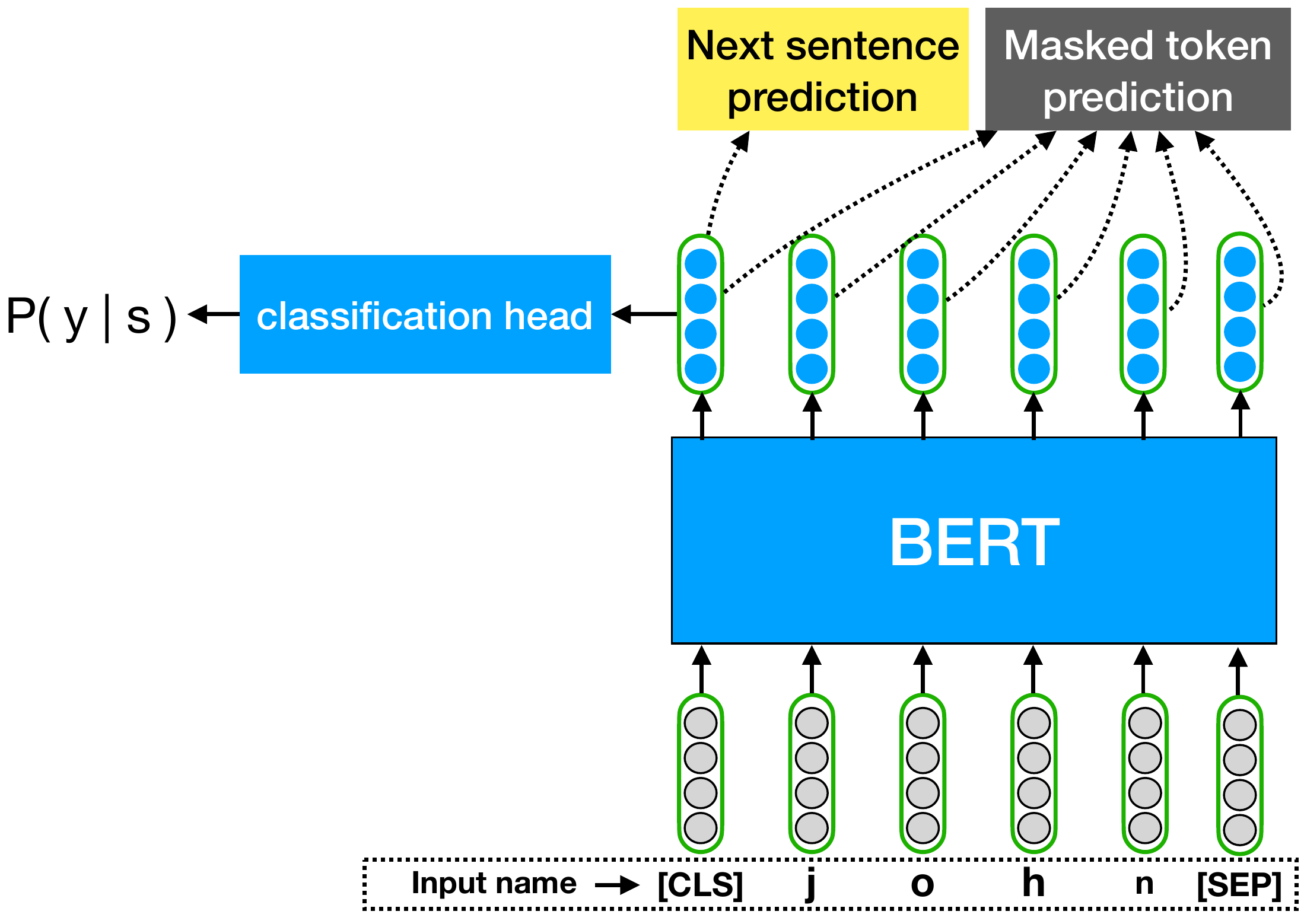}
		\caption{Traditional Fine-tuning BERT.}
		\label{fig:bert-traditional}
	\end{subfigure}
	\hfill
	\begin{subfigure}{0.58\textwidth}
		\centering
		\includegraphics[width=\textwidth]{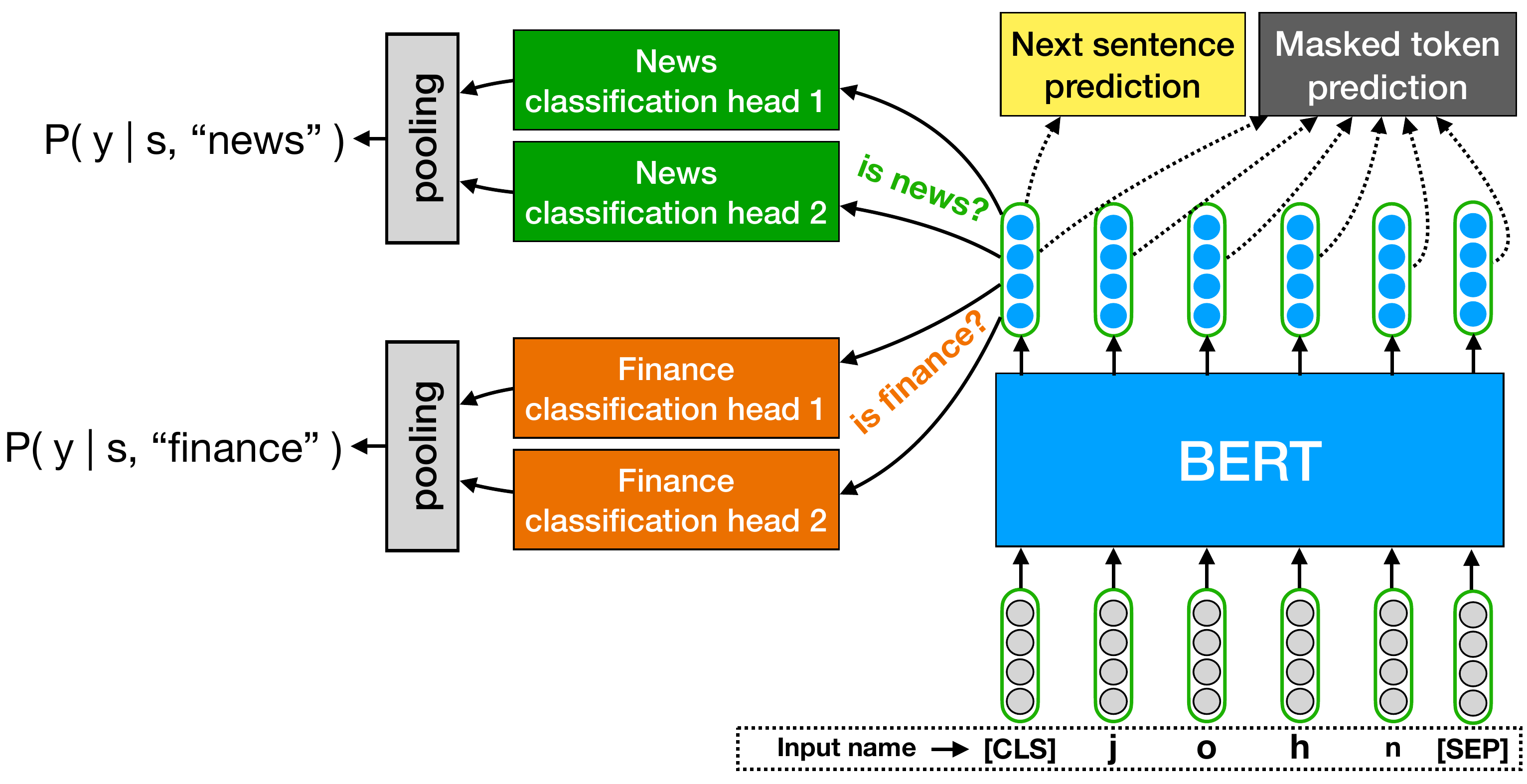}
		\caption{HABERTOR with two sources and ensemble of 2 heads.}
		\label{fig:bert-flex-ensemble}
	\end{subfigure}
    \vspace{-10pt}
    \caption{Architecture comparison of traditional fine-tuned BERT and HABERTOR multi-source ensemble heads.}
    \label{fig:architecture}
    \vspace{-15pt}
\end{figure*}




\subsection{Tokenization} BERT model relies on WordPiece (WP)~\cite{wu2016google}, a Google's internal code that breaks down each word into common sub-word units (``wordpieces'').
These sub-words are like character n-grams, except that they are automatically chosen to ensure that
each of these sub-words is frequently observed in the input corpus.
WP improves handling of rare words, such as intentionally mis-spelled abusive words, without the need for a huge vocabulary.  A comparable implementation that is open sourced is SentencePiece (SP)~\cite{kudo2018sentencepiece}. Like WP, the vocab size is predetermined. Both WP and SP are unsupervised learning models.
Since WP is not released in public, we train a SP model using our training data, then use it to tokenize input texts.

\subsection{Parameter Reduction with Quaternion Factorization}
Denote V  the vocab size, E  the embedding size, H  the hidden size, L  the number of layers, and F the feed-forward/filter size. In BERT, F = 4H, E = H, and the number of attention heads is $H/64$. 
Encoding the vocabs takes VH parameters.  
Each BERT layer contains three parts: (i) attention, (ii) filtering/feedforward, and (iii) output. Each of the three parts has 4$\text{H}^2$ parameters. 
Thus, a BERT layer has 12$\text{H}^2$ parameters and a BERT-base setting with 12 layers has $\text{VH} + 144 \text{H}^2$ parameters. 
Please refer to Section \ref{sec:bert-param-analysis} in the Appendix for details.


Recently, Quaternion representations have shown its benefits over Euclidean representations in many neural designs \cite{parcollet2018quaternion,tay-etal-2019-lightweight}: (i) a Quaternion number consists of a real component and three imaginary components, encouraging a richer extent of expressiveness; and (ii) a Quaternion transformation reduces 75\% parameters compared to the traditional Euclidean transformation because of the weight sharing using the Hamilton product. Hence, we propose Quaternion fatorization strategies to significantly reduce the model's parameters as follows:

\noindent\textbf{Vocab Factorization (VF)}: Inspired by \citet{Lan2020ALBERT}, we encode V vocabs using Quaternion representations with an embedding size E$\ll$H. Then, we apply a Quaternion transformation to transform E back to H, and concatenate all four parts of a Quaternion to form a regular Euclidean embedding. This leads to a total of VE + EH/4 parameters, compared to VE + EH in ALBERT.

\noindent\textbf{Attention Factorization (AF)}: If the input sequences have length N, the output of the multi-head attention is N$\times$N, which does not depend on the hidden size H. Hence, it is unnecessary to produce the attention Query, Key, and Value with the same input hidden size H and cost 3$\text{H}^2$ parameters per a layer. Instead, we produce the attention Query, Key, and Value in size C$\ll$H using linear Quaternion transformations, leading to 3CH/4 parameters.

\noindent\textbf{Feedforward Factorization (FF):} Instead of linearly transforming from H to 4H (i.e. 4$H^2$ parameters), we apply Quaternion transformations from H to I, and from I to 4H, with I$\ll$H is an intermediate size. This leads to a total of (HI/4 + IH) parameters.

\noindent\textbf{Output Factorization (OF):} We also apply Quaternion transformations from 4H to I, then from I to H. This results in (HI + IH/4) parameters, compared to 4$H^2$ in BERT.

When we apply all the above compression techniques together, the total parameters are reduced to VE + EH/4 + L(3CH/4 + $H^2$ + 5HI/2). Particularly, with BERT-base settings of V=32k, H=768, L=12, if we set E=128, C=192, and I=128, the total of parameters is reduced from 110M to \textbf{only 8.4M}.

\subsection{Pretraining tasks}
Similar to BERT, we pre-train our {HABERTOR} 
with two unsupervised learning/\emph{language modeling} tasks: (i) masked token prediction, and (ii) next sentence prediction. 
We describe some modifications that we made to the original BERT's implementation as follows:

\subsubsection{Masked token prediction task}
BERT generates only one masked training instance for each input sequence. Instead, inspired by \citet{liu2019roberta}, we generate $\tau$ training instances by randomly sampling with replacement masked positions $\tau$ times. We refer to $\tau$ as a \emph{masking factor}. Intuitively, this helps the model to learn differently combined patterns of tokens in the same input sequence, and boosts the model's language understanding. This small modification works especially well when we have a smaller pre-training data size, which is often true for a domain-specific task (e.g., hatespeech detection).

\subsubsection{Next sentence prediction task}
In BERT, the two input sentences are already paired and prepared in advanced. In our case, we have to preprocess input text sequences to prepare paired sentences for the next sentence prediction task. We conduct the following preprocessing steps:

\textbf{Step 1:} We train an unsupervised sentence tokenizer from \emph{nltk} library. Then we use the trained sentence tokenizer to tokenize each input text sequence into (splitted) sentences.

\textbf{Step 2:} 
In BERT, 50\% of the chance two consecutive sentences are paired as \emph{next}, and 50\% of the chance two non-consecutive sentences are paired as \emph{not next}. In our case, our input text sequences can be broken into one, two, three, or more sentences. For input text sequences that consist of only one tokenized sentence, the only choice is to pair with another random sentence to generate a \emph{not next} example. By following our 50-50 rule described in the Appendix, we ensure generating an equal number of \emph{next} and \emph{not next} examples. 

\vspace{-3pt}
\subsection{Training the hatespeech prediction task}
For hatespeech prediction task, 
we propose a multi-source multi-head {HABERTOR} classifier. 
The architecture comparison of the traditional fine-tuning BERT and our proposal is shown in Figure \ref{fig:architecture}. We note two main differences in our design as follows.

First, as shown in Figure \ref{fig:bert-flex-ensemble}, our {HABERTOR} has separated classification heads/nets for different input sequences of different sources but with a shared language understanding knowledge.
Intuitively, instead of measuring the same probabilities $P (y| s)$ for all input sequences, it injects additional prior source knowledge of the input sequences to measure $P (y| s,$ ``news'') or $P (y| s$, ``finance''). 

Second, in addition to multi-source, {HABERTOR} with an \emph{ensemble} of $h$ heads provides even more capabilities to model data variance. For each input source, we employ ensemble of several classification heads (i.e. two classification heads for each source in the Figure \ref{fig:bert-flex-ensemble}) and use a pooling layer on top to aggregate results from those classification heads. We use three pooling functions: \emph{min, max, mean}. \emph{min} pooling indicates that {HABERTOR} classifies an input comment as a hateful one if all of the heads classify it as hatespeech, which put a more stringent requirement on classifying hatespeech. On the other hand, {HABERTOR} will predict an input comment as a normal comment if at least one of the heads recognizes the input comment as a normal one, which is less strict. Similarly, using \emph{max} pooling will put more restriction on declaring
comments as normal, and less restriction on declaring hatespeech. Finally, \emph{mean} pooling considers the average voting from all heads.

Note that our design generalizes the traditional fine-tuning BERT architecture when $h$=1 and the two classification nets share the same weights. Thus, {HABERTOR} is more flexible than the conventional fine-tuning BERT.
Also, {HABERTOR} can be extended trivially to problems that have $q$ sources, with $h$ separated classification heads for $q$ different sources. When predicting input sequences from new sources, {HABERTOR} averages the scores from all separated classification nets.


\vspace{-3pt}
\subsection{Parameter Estimation}

Estimating parameters in the pretraining tasks in our model is similar to BERT, and we leave the details in the Appendix due to space limitation. 

For hatespeech prediction task, we use the transformed embedding vector of the \emph{[CLS]} token as a summarized embedding vector for the whole input sequence. Let $\boldsymbol{S}$ be a collection of sequences $s_i$. 
Note that $s_i$ is a normal sequence, not corrupted or concatenated with another input sequence. Given that $\boldsymbol{y}_i$ is the supervised ground truth label for the input sequence, and $\boldsymbol{\hat{y}}_i=P(\boldsymbol{y}_i| s_i,$ ``news'') (Figure \ref{fig:bert-flex-ensemble}, \ref{fig:bert-flex-ensemble}) where $s_i$ is a \emph{news} input sequence, or $\boldsymbol{\hat{y}}_i=P(\boldsymbol{y}_i | s_i,$``finance'') when $s_i$ is a \emph{finance} input sequence. The hateful prediction task aims to minimize the following binary cross entropy loss:
\vspace{-0.1cm}
\begin{equation}
\label{equa:hate-speech-loss}
\nonumber
\resizebox{0.5\textwidth}{!}{$
\begin{aligned}
\mathcal{L}_{hs} = \operatorname*{argmin}_{\theta} \;
-\sum_{i=1}^{|\boldsymbol{S}|}
& \boldsymbol{y}_i \log   \big( \boldsymbol{\hat{y}}_i\big) +
 (1 - \boldsymbol{y}_i) \log \; \big( 1 - \boldsymbol{\hat{y}}_i \big)
\end{aligned}
$}
\vspace{-0.1cm}
\end{equation}


\noindent\textbf{Regularize with adversarial training:} To make our model more robust to perturbations of the input embeddings, we further regularize our model with adversarial training. There exist several state-of-the-art \emph{target}-based adversarial attacks such us Fast Gradient Method (\emph{FGM}) \cite{miyato2016adversarial}, Basic Iterative Method \cite{kurakin2016adversarial}, and Carlini L2 attack \cite{carlini2017towards}. We use the \emph{FGM} method as it is effective and efficient according to our experiments. 

In \emph{FGM}, the noise magnitude is a scalar value and is a manual input hyper-parameter. This is sub-optimal, as different adversarial directions of different dimensions are scaled similarly, plus, manually tuning the noise magnitude is expensive and not optimal. 
Hence, we propose to extend \emph{FGM} with a \emph{learnable} and \emph{fine-grained} noise magnitude, where the noise magnitude is parameterized by a learnable vector, providing different scales for different adversarial dimensions. Moreover, the running time of our proposal compared to \emph{FGM} is similar. 

The basic idea of the adversarial training is to add a small perturbation noise $\delta$ on each of the token embeddings that makes the model mis-classify hateful comments as normal comments, and vice versa. Given the input sequence $s_i=[w_1^{(i)}, w_2^{(i)}, ..., w_u^{(i)}]$ with ground truth label $y_i$, let $\Tilde{y}_i$ be the adversarial target class of $s_i$ such that $\Tilde{y}_i \ne y_i$. In the hatespeech detection domain, our model is a binary classifier. Hence, when $y_i = 1$ ($s_i$ is a hateful comment), $\Tilde{y}_i = 0$ and vice versa. Then, the perturbation noise $\delta$ is learned by minimizing the following cost function:
\vspace{-0.1cm}
\begin{equation}
\label{equa:hate-speech-adv-loss1}
\resizebox{0.42\textwidth}{!}{$
\begin{aligned}
& \mathcal{L}_{adv} = \operatorname*{argmin}_{\delta, \delta \in [a, b]} \;
-\sum_{i=1}^{|\boldsymbol{S}|}
log P
\big(
  \boldsymbol{\Tilde{y}_i} \; | s_i + \delta_i ; \hat{\theta}
\big)
\end{aligned}
$}
\vspace{-0.05cm}
\end{equation}

Note that in Eq. (\ref{equa:hate-speech-adv-loss1}), $\delta$ is constrained to be less than a predefined noise magnitude scalar in the traditional FGM method. In our proposal, $\delta$ is constrained within a range $[a, b]$ (i.e. $min(\delta) \ge$ a $\wedge$ $max(\delta) \le$ b ). 
\noindent Solving Eq. (\ref{equa:hate-speech-adv-loss1}) is expensive and not easy, especially with complicated deep neural networks. Thus, we approximate each perturbation noise $\delta_i$ for each input sequence $s_i$ by linearizing partial loss $- log P\big(\boldsymbol{\Tilde{y}_i} \; | s_i + \delta_i ; \hat{\theta} \big)$ around $s_i$.
Particularly, $\delta_i$ is measured by:
\vspace{-0.1cm}
\begin{equation}
\label{equa:hate-speech-adv-loss2}
\resizebox{0.35\textwidth}{!}{$
\begin{aligned}
\delta_i = -\epsilon \times
         \frac {\bigtriangledown_{s_i} \big( -log P\big(\boldsymbol{\Tilde{y}_i} \; | s_i ; \hat{\theta} \big) \big) }
               {\Vert \bigtriangledown_{s_i} \big(-log P\big(\boldsymbol{\Tilde{y}_i} \; | s_i ; \hat{\theta} \big) \big) \Vert_2}
\end{aligned}
$}
\vspace{-0.05cm}
\end{equation}
In Eq.~(\ref{equa:hate-speech-adv-loss2}), $\epsilon$ is a {\em learnable vector}, with the same dimensional size as $\delta_i$. Solving the constraint $\delta_i \in [a, b]$ in Eq.~(\ref{equa:hate-speech-adv-loss1}) becomes restricting $\epsilon \in [a, b]$, which is trivial by projecting $\epsilon$ in $[a, b]$.

Finally, {HABERTOR} aims to minimize the following cost function:
\vspace{-0.1cm}
\begin{equation}
\label{equa:hate-speech-final-loss}
\mathcal{L} = \mathcal{L}_{hs} + \lambda_{adv}\mathcal{L}_{adv} -  \lambda_{\epsilon} \Vert \epsilon \Vert_2,
\vspace{-0.1cm}
\end{equation}
\noindent where $\Vert \epsilon \Vert_2$ is an additional term to force the model to learn robustly as much as possible, and $\lambda_{\epsilon}$ is a hyper-parameter to balance its effect. Note that, we first learn the optimal values of all token embeddings and {HABERTOR}'s parameters before learning adversarial noise $\delta$. Also, regularizing adversarial training only increases the training time, but not the inferencing time since it does not introduce extra parameters for the model during the inference.

%% file: 5-experiments.tex
\section{Empirical Study}
\label{sec:emperical}

\subsection{Experiment Setting}
\noindent\textbf{Dataset:}
Our primary dataset was extracted from user comments on Yahoo News and Finance for five years, and consisted of 1,429,138 labeled comments. Among them, 944,391 comments are from Yahoo News and 484,747 comments are from Yahoo Finance. There are 100,652 hateful comments. 
The 1.4M labeled data was collected as follows~\cite{hate_speech_Nobata_2016}: comments that are reported as ``abusive'' for any reason by users of Yahoo News and Finance are sent to in-house trained raters for review and labeling. 

To further validate the generalizability of {HABERTOR}, we perform transfer-learning experiments on other two publicly available hatespeech datasets:
Twitter \cite{waseem2016hateful}, and Wikipedia (i.e. Wiki) \cite{wulczyn2017ex}.
The Twitter dataset consists of 16K annotated tweets, including 5,054 hateful tweets (i.e., 31\%). The Wiki dataset has 115K labeled discussion comments from English Wikipedia talk's page, including 13,590 hateful comments (i.e., 12\%). The statistics of 3 datasets are shown in Table~\ref{table:datasets}.


\noindent\textbf{Train/Dev/Test split:} We split the dataset into train/dev/test sets with a ratio 70\%/10\%/20\%. 
We tune hyper-parameters on the dev set, and report final results on the test set. Considering critical mistakes reported at \citet{arango2019hate} when building machine learning models (i.e. extracting features using the entire dataset, including testing data, \emph{etc}), we generate vocabs, pre-train the two language modeling tasks, and train the hatespeech prediction task using \textbf{only the training set}.

\begin{table}[t]
    \centering
\setlength{\tabcolsep}{2pt}
\caption{Statistics of the three datasets.}
     \vspace{-10pt}
    \label{table:datasets}
    \resizebox{0.75\linewidth}{!}{
        \begin{tabular}{lcccc}
            \toprule
            Statistics/Datasets & Yahoo & Twitter & Wiki \\
            \midrule
            Total               & 1.4M  & 16K   & 115K \\
            \# Hateful          & 100K  & 5K    & 13K \\
            \% of hatespeech    & 7\%   & 31\%  & 12\% \\

            \bottomrule
        \end{tabular}
    }
    \vspace{-12pt}
\end{table}

\noindent\textbf{Baselines, our Models and hyper-parameter Settings:}
We compare our models with 15 state-of-the-art baselines: Bag of Words (BOW) \cite{dinakar2011modeling,van2015automatic}, NGRAM, CNN \cite{kim2014convolutional}, VDCNN \cite{conneau2016very}, FastText \cite{joulin2016fasttext}, LSTM \cite{cho2014learning}, att-LSTM, RCNN \cite{lai2015recurrent}, att-BiLSTM \cite{lin2017structured}, Fermi (best hatespeech detection method as reported in \citet{basile-etal-2019-semeval}) \cite{indurthi-etal-2019-fermi}, Q-Transformer \cite{tay-etal-2019-lightweight}, Tiny-BERT \cite{jiao2019tinybert}, DistilBERT-base \cite{sanh2019distilbert}, ALBERT-base \cite{Lan2020ALBERT}, and BERT-base \cite{devlin2018bert,nikolov2019nikolov}. We are aware of other recent language models such as Transformer-XL \cite{dai2019transformer}, RoBERTa\cite{liu2019roberta}, DialoGPT \cite{zhang2019dialogpt}, to name a few. However, as these models are even heavier than BERT-base, we do not compare with them. The detailed description of the baselines and hyper-parameter settings is described in the Appendix.

\noindent\textbf{Our models:} We denote \emph{HABERTOR} as our model without using any factorization, \emph{HABERTOR-VQF} as \emph{HABERTOR} + Vocab Quaternion Factorization, \emph{HABERTOR-VAQF} as \emph{HABERTOR} + Vocab + Attention Quaternion Factorization, \emph{HABERTOR-VAFQF} as \emph{HABERTOR} + Vocab + Attention + Feedforward Quaternion Factorization, and \emph{HABERTOR-}\emph{VAFOQF} as \emph{HABERTOR} + Vocab + Attention + Feedforward + Output Quaternion Factorization.



\noindent\textbf{Measurement:}
We evaluate models on seven metrics: Area Under the Curve (\emph{AUC}), Average Precision (\emph{AP}), False Positive Rate (FPR),  False Negative Rate (FNR), F1 score\footnote{Both AP and F1 account for Precision and Recall so we do not further report Precision and Recall for saving space.}. In real world, for imbalanced datasets, we care more about FPR and FNR. Thus, we report FPR at 5\% of FNR (FPR@5\%FNR), meaning we allow 5\% of hateful texts to be misclassified as normal ones, then report FPR at that point. Similarly, we report FNR at 5\% of FPR (FNR@5\%FPR).
Except for AUC and AP, the other metrics are reported using an optimal threshold 
selected by using the development set.

\begin{table}[t]
	\centering
	\setlength{\tabcolsep}{2pt}
	\caption{Parameters Comparison between \emph{HABERTOR-VAFOQF} vs. other LMs. ``--'' indicates not available.}
	\vspace{-10pt}
	\label{table:config}
	\resizebox{1.\linewidth}{!}{
		\begin{tabular}{lcccccc}
			\toprule
			Statistics        & \specialcell{\textbf{HABERTOR}\\-\textbf{VAFOQF}} & 
			\specialcell{AL-\\BERT} &
			\specialcell{Tiny-\\BERT} & \specialcell{Distil-\\BERT} & \specialcell{BERT-\\base} \\
			\midrule
			Layers (L)         & \textbf{6}        & 12      & 4     & 6     & 12  \\
			Attention heads    & \textbf{6}        & 12      & 12    & 12    & 12  \\
			Attention size (C) & \bf{192}          & --     & --   & --   & -- \\       
			
			Embedding (E)      & \textbf{128}      & 128     & --   & --   & -- \\
			Hidden (H)         & \textbf{384}      & 768     & 312   & 768   & 768 \\
			Intermediate size (I)& \textbf{128}    & --     &  --  & --   & -- \\
			Feedforward size     & \textbf{1,536}    & 3072    & 1,200 & 3,072 & 3,072 \\
			Vocab (V)          & \textbf{40k}      & 30k     & 30k   & 30k   & 30k \\
			
			\midrule
			Parameters        & \bf{7.1M}   & 12M & 14.5M & 65M &110M \\
			\bottomrule
		\end{tabular}
	}
	\vspace{-15pt}
\end{table}
\begin{table*}[t]
	\centering
	\caption{Performance of all models that we train on Yahoo train data, test on Yahoo test data and report results on Yahoo News and Yahoo Finance separately. 
		Best baseline is \underline{underlined}, \textbf{better} results than best baseline are \textbf{bold}.}
	\vspace{-8pt}
	\label{table:yahoo-performance}
	\resizebox{\textwidth}{!}{
		\begin{tabular}{l|cc|ccccc|ccccc}
			\toprule
			\multicolumn{1}{c}{\multirow{2}{*}{Model}} &
			\multicolumn{2}{|c}{\bf Yahoo} &
			\multicolumn{5}{|c}{\bf Yahoo News} & \multicolumn{5}{|c}{\bf Yahoo Finance}                                \\
			\multicolumn{1}{c}{} &
			\multicolumn{1}{|c}{AUC} & \multicolumn{1}{c}{AP} &
			\multicolumn{1}{|c}{AUC} & \multicolumn{1}{c}{AP} &  \multicolumn{1}{c}{\begin{tabular}[c]{@{}c@{}}FPR@\\ 5\%FNR\end{tabular}} & \begin{tabular}[c]{@{}c@{}}FNR@\\ 5\%FPR\end{tabular} & F1 & \multicolumn{1}{|c}{AUC} & \multicolumn{1}{c}{AP} & \multicolumn{1}{c}{\begin{tabular}[c]{@{}c@{}}FPR@\\ 5\%FNR\end{tabular}} & \begin{tabular}[c]{@{}c@{}}FNR@\\ 5\%FPR\end{tabular} & F1 \\
			\midrule
			%
			BOW        & 85.91 & 48.35
			& 85.07 & 51.37 & 61.13 & 50.53 & 49.01
			& 85.83 & 36.80 & 60.97 & 49.43 & 40.15\\
			NGRAM      & 84.19 & 42.15
			& 83.17 & 45.00 & 63.45 & 57.45 & 43.59
			& 84.29 & 31.63 & 63.42 & 53.94 & 35.95\\
			CNN        & 91.21 & 63.03
			& 90.64 & 65.64 & 47.50 & 36.23 & 60.61
			& 91.20 & 52.30 & 45.59 & 33.96 & 51.93 \\
			VDCNN      & 88.10 & 58.08
			& 87.65 & 60.75 & 60.39 & 41.56 & 56.12
			& 88.17 & 48.72 & 62.43 & 38.78 & 50.38 \\
			FastText   & 91.64 & 60.15
			& 90.97 & 63.16 & 41.80 & 38.09 & 58.35
			& 92.13 & 47.97 & 37.75 & 34.30 & 49.36 \\
			LSTM       & 91.83 & 64.17
			& 91.14 & 66.59 & 43.81 & 35.09 & 60.96
			& 92.38 & 54.44 & 38.26 & 31.45 & 53.36 \\
			att-LSTM   & 91.83 & 64.39
			& 91.10 & 66.77 & 44.24 & 34.86 & 61.37
			& 92.43 & 54.79 & 38.32 & 30.75 & 53.79 \\
			RCNN       & 91.17 & 63.34
			& 90.52 & 65.72 & 48.49 & 36.37 & 60.29
			& 91.32 & 53.77 & 49.40 & 32.17 & 52.73 \\
			att-BiLSTM & 92.52 & 64.17
			& 91.93 & 66.82 & 38.07 & 34.68 & 61.54
			& 92.93 & 53.97 & 36.05 & 31.14 & 52.58 \\
			Fermi      & 86.53 & 41.52
			& 86.10 & 45.16 & 53.33 & 55.60 & 45.65
			& 85.45 & 27.53 & 56.60 & 56.48 & 33.27 \\
			Q-Transformer
			& 92.34 & 64.43
			& 91.81 & 67.06 & 39.12 & 34.17 & 61.82
			& 92.64 & 54.41 & 37.71 & 29.74 & 53.51\\
			Tiny-BERT  & 93.60 & 68.70
			& 93.03 & 70.80 & 34.50 & 30.37 & 64.42
			& 94.09 & 60.25 & 31.16 & 25.09 & 57.58 \\
			DistilBERT & 93.68 & 69.15
			& 93.13 & 71.25 & 34.33 & 30.05 & 64.69
			& 94.12 & 60.56 & 29.23 & 24.94 & 58.01 \\
			ALBERT     & 93.50 & 67.99
			& 92.93 & 70.28 & 34.56 & 31.15 & 63.82
			& 93.94 & 58.73 & 30.12 & 25.87 & 56.37 \\
			BERT-base  & \underline{94.14} & \underline{70.05}
			& \underline{93.56} & \underline{71.65} & \underline{32.15} & \underline{28.91} & \underline{65.30}
			& \underline{94.60} & \underline{62.34} & \underline{29.14} & \underline{22.81} & \underline{59.72} \\
			
			\midrule
			HABERTOR
			& {\bf 94.77} & {\bf 72.35}
			& {\bf94.12} & {\bf73.79} & {\bf29.26} & {\bf27.12} & {\bf67.09}
			& {\bf95.72} & {\bf65.93} & {\bf22.03} & {\bf18.99} & {\bf62.38} \\
			HABERTOR-VQF
			& \textbf{94.70}	& \textbf{71.82}	
			& \textbf{94.00}	& \textbf{73.25}	& \textbf{29.50	}& \textbf{27.79}	& \textbf{66.57	}
			& \textbf{95.81}	& \textbf{65.20}	& \textbf{20.78}	& \textbf{20.08	}& \textbf{61.60}
			\\
			HABERTOR-VAQF
			&  \textbf{94.59} & \textbf{71.53}
			&  \textbf{93.90} & \textbf{73.02} & \textbf{29.94} & \textbf{27.92} & \textbf{66.51}
			&  \textbf{95.63} & \textbf{64.64} & \textbf{23.08} & \textbf{20.39} & \textbf{60.84} \\
			HABERTOR-VAFQF
			& \textbf{94.43} & \textbf{70.75}
			& \textbf{93.72} & \textbf{72.37} & \textbf{31.86} & \textbf{28.58} & \textbf{65.81}
			& \textbf{95.42} & \textbf{63.07} & \textbf{22.87} & \textbf{21.43} & \textbf{60.11}\\
			HABERTOR-VAFOQF
			& \textbf{94.18} & 69.92
			& 93.51 & 71.63 & 32.47 & 29.26 & \textbf{65.35}
			& \textbf{95.00} & 61.99 & \textbf{24.95} & 22.81 & 59.50 \\

			
			
		\end{tabular}
	}
	\vspace{-12pt}
\end{table*}

\noindent \textbf{Model size comparison}: 
\emph{HABERTOR} has 26M of parameters. 
\emph{HABERTOR-VQF} and \emph{HABERTOR-VAQF} have 16.2M and 13.4M of parameters, respectively. 
\emph{HABERTOR-VAFQF} and \emph{HABERTOR-VAFOQF} have 10.3M and 7.1M of parameters, respectively. 
The size of all five models is much smaller than BERT-base (i.e. 110M of parameters). 
The configuration comparison of {HABERTOR-VAFOQF} and other pretrained language models 
is given in Table~\ref{table:config}. 
\emph{HABERTOR-VAFOQF} has less than 2 times compared to TinyBERT's parameters, less than 9 times compared to Distil-BERT's size, and is equal to 0.59 AlBERT's size. 

\subsection{Experimental results}

\subsubsection{Performance comparison}
Table~\ref{table:yahoo-performance} shows the performance of all models on Yahoo dataset. Note that we train on the Yahoo training set that contains both Yahoo News and Finance data, and report results on Yahoo News and Finance separately, and report only AUC and AP on both of them (denoted as column ``Yahoo'' in Table~\ref{table:yahoo-performance}). We see that {Fermi} worked worst among all models. It is mainly because Fermi transfers the pre-trained embeddings from the USE model to a SVM classifier without further fine-tuning. This limits {Fermi}'s ability to understand domain-specific contexts. 
Q-Transformer works the best among non-LM baselines, but worse than LM baselines as it is not pretrained. 
{BERT-base} performed the best among all baselines. 
Also, distilled models worked worse than {BERT-base} due to their compression nature on BERT-base as the teacher model.


Next, we compare the performance of our proposed models against each other. Table \ref{table:yahoo-performance} shows that our models' performance is decreasing when we compress more components (\emph{p-value} $<$ 0.05 under the directional Wilcoxon signed-rank test). We reason it is a trade-off between the model size and the model performance as factorizing a component will naturally lose some of its information. 

Then, we compared our proposed models with BERT-base -- the best baseline. Table \ref{table:yahoo-performance} shows that except our \emph{HABERTOR-VAFOQF}, 
our other proposals outperformed BERT-base, improving by an average of 1.2\% and 1.5\% of F1-score in Yahoo News and Yahoo Finance, respectively (\emph{p-value} $<$ 0.05). Recall that in addition to improving hatespeech detection performance, our models' size is much smaller than BERT-base. For example, \emph{HABERTOR} saved 84M of parameters from BERT-base, and \emph{HABERTOR-VAFQF} saved nearly 100M of parameters from BERT-base. Interestingly, even our smallest \emph{HABERTOR-VAFOQF} model (7.1M of parameters) achieves similar results compared to BERT-base (i.e. the performance difference between them is not significant under the directional Wilcoxon signed-rank test). 
Those results show the effectiveness of our proposed models against {BERT-base}, the best baseline, and consolidate the need of pretraining a language model on a hateful corpus for a better hateful language understanding.
\subsubsection{Running time and memory comparison}
\noindent\textbf{Running time:} Among LM baselines, TinyBERT is the fastest. Though ALBERT has the smallest number of parameters by adopting the cross-layer weight sharing mechanism, ALBERT has the same number of layers as BERT-base, leading to a similar computational expense as BERT-base.

Our \emph{HABERTOR-VQF} and \emph{HABERTOR-VAQF} have a very similar parameter size with \emph{TinyBERT} and their train/inference time are similar. Interestingly, even though \emph{HABERTOR} has 26M of parameters, its runtime is also competitive with \emph{TinyBERT}. This is because among 26M of parameters in \emph{HABERTOR}, 15.4M of its total parameters are for encoding 40k vocabs, which are not computational parameters and are only updated sparsely during training. \emph{HABERTOR-VAFQF} and \emph{HABERTOR-VAFOQF} significantly reduce the number of parameters compared to TinyBERT, leading to a speedup during training and inference phases. Especially, our experiments on 4 K80 GPUs with a batch size of 128 shows that \emph{HABERTOR-VAFOQF} is 1.6 times faster than TinyBERT.

\noindent\textbf{Memory consumption:} Our experiments with a batch size of 128 on 4 K80 GPUs show that among LM baselines, TinyBERT and ALBERT are the most lightweight models, consuming ~13GB of GPU memory. Compared to TinyBERT and ALBERT, \emph{HABERTOR} takes an additional 4GB of GPU memory, while \emph{HABERTOR-VQF}, \emph{HABERTOR-VAQF} hold a similar memory consumption, \emph{HABERTOR-VAFQF} and \emph{HABERTOR-VAFOQF} reduces 1$\sim$3 GB of GPU memory.

\noindent\textbf{Compared to BERT-base:} In general, \emph{HABERTOR} is 4$\sim$5 times faster, and 3.1 times GPU memory usage smaller than {BERT-base}. Our most lightweight model \emph{HABERTOR-VAFOQF} even reduces 3.6 times GPU memory usage, while remains as effective as BERT-base. The memory saving in our models also indicates that we could increase the batch size to perform inference even faster.

\begin{table}
\centering
\caption{Generalizability of \emph{HABERTOR} and top baselines. Report AUC, AP, and F1 on each test set.}
\vspace{-10pt}
\label{table:generalizability}
\resizebox{0.5\textwidth}{!}{
\begin{tabular}{l|lll|lll}
\toprule
      & \multicolumn{3}{c|}{Twitter}                            & \multicolumn{3}{c}{Wiki} \\
      \toprule
Model & \multicolumn{1}{c}{AUC} & \multicolumn{1}{c}{AP} & F1 & \multicolumn{1}{c}{AUC} & \multicolumn{1}{c}{AP} & F1 \\
      \midrule
Fermi & 89.03 &	79.23 &	74.52&	96.59&	84.26&	75.51\\
TinyBERT & 92.23 & 83.88 & 78.33 & 97.10 & 87.64 & 79.70 \\
DistilBERT & 92.13 & 80.21 & 77.89 & 97.23 & 88.16 & 80.21 \\
ALBERT & 92.55 &	86.51	& 78.76 & 97.66 &	88.91 &	80.66 \\
BERT & \underline{93.21} & \underline{86.67} & \underline{79.68} & \underline{97.75} & \underline{89.23} & \underline{80.73} \\
\midrule
HABERTOR        &
        \textbf{93.52} & \textbf{88.57} & \textbf{81.22 } &
        97.46 & 88.65 & \textbf{80.81} \\
HABERTOR-VQF  &
        \textbf{93.94} &	\textbf{88.45} &	\textbf{81.21} &
        97.40 & 88.64 & 80.66 \\
HABERTOR-VAQF   &
        \textbf{93.57}	& \textbf{87.66}	& \textbf{80.23} &
        97.45	& 88.61 & 80.63 \\
HABERTOR-VAFQF  &
        \textbf{93.51}	& \textbf{87.38} & \textbf{80.16}	&
        97.37	& 88.21	& 80.23 \\
HABERTOR-VAFOQF   &
        \textbf{93.49} & \textbf{87.14} & \textbf{80.06}	&
        97.23	& 87.94 & 79.61 \\
\bottomrule
\end{tabular}
}
\vspace{-15pt}
\end{table}

\subsubsection{Generalizability analysis}
We perform hatespeech Language Model transfer learning on other hateful Twitter and Wiki datasets to understand our models' generalizability. We use our models' pre-trained language model checkpoint learned from Yahoo hateful datasets, and fine tune them on Twitter/Wiki datasets. Note that the fine-tuned training also includes regularized adversarial training for best performance. 
Next, we compare the performance of our models with Fermi and four LM baselines -- best baselines reported in Table \ref{table:yahoo-performance}.  

Table~\ref{table:generalizability} shows that {BERT-base} performed best compared to other fine-tuned LMs, which is consistent with our reported results on Yahoo datasets in Table \ref{table:yahoo-performance}. When comparing with BERT-base's performance (i.e. best baseline) on the Twitter dataset, all our models outperformed BERT-base. 
On Wiki dataset, interestingly, our models work very competitively with BERT-base, and achieve similar F1-score results. Recall that BERT-base has a major advantage of pre-training on 2,500M Wiki words, thus potentially understands Wiki language styles and contexts better. In contrast, HABERTOR and its four factorized versions are pre-trained on 33M words from Yahoo hatespeech dataset. As shown in the ablation study (refer to \emph{AS2} in Section \ref{sec:ablation} of the Appendix), a larger pre-training data size leads to better language understanding and a higher hatespeech prediction performance. 
Hence, if we acquire larger pre-training data with more hateful representatives, our model's performance can be further boosted. 
All of those results show that our models generalize well on other hatespeech datasets compared with BERT-base, with significant model complexity reduction.

\vspace{-3pt}
\subsubsection{Ablation study}
\begin{table}[]
	\centering
	\caption{Comparison of the \textbf{traditional} FGM with a \textit{fixed} and \textit{scalar} noise magnitude, compared to the FGM with \textbf{our} proposed \textit{fine-grained} and \textit{adaptive} noise magnitude. Better
		results are in \textbf{bold}.}
	\vspace{-5pt}
	\label{table:adv-comparison}
	\resizebox{0.5\textwidth}{!}{
		\begin{tabular}{l|l|lll|lll}
			\toprule
			\multicolumn{1}{l|}{}                             & \multicolumn{1}{l|}{}                 & \multicolumn{3}{c|}{Twitter}                          & \multicolumn{3}{c}{Wiki}                             \\ \hline
			\multicolumn{1}{c|}{Model}                       & \multicolumn{1}{c|}{Type}             & \multicolumn{1}{c}{AUC} & \multicolumn{1}{c}{AP} & F1 & \multicolumn{1}{c}{AUC} & \multicolumn{1}{c}{AP} & F1 \\ 
			\midrule
			{\multirow{2}{*}{HABERTOR}} 
			& {traditional}    
			& 93.54 & 87.88 & 79.84 
			& 97.50 & 88.14 & 80.13
			\\
			\cline{2-8}
			& {\textbf{ours}}         
			& 93.52 & \textbf{88.57} & \textbf{81.22} 
			& 97.46 & \textbf{88.65} & \textbf{80.81} 
			\\
			\midrule
			\midrule
			{\multirow{2}{*}{\specialcell{HABERTOR-\\VQF}}} 
			& {traditional}         
			& 93.62 & 88.09 & 80.26 
			& 97.44 & 88.19 & 80.11 \\
			\cline{2-8}
			& {\textbf{ours}}         
			& 93.94 & \textbf{88.45} & \textbf{81.21}
			& 97.40 & \textbf{88.64} & \textbf{80.66} \\
			\midrule
			\midrule
			{\multirow{2}{*}{\specialcell{HABERTOR-\\VAQF}}}
			& {traditional}         
			& 93.03 & 86.77 & 79.56 
			& 97.44 & 88.15 & 80.16 \\
			\cline{2-8}
			& {\textbf{ours}}         
			& 93.57	& \textbf{87.66}	& \textbf{80.23}
			& 97.45	& \textbf{88.61} & \textbf{80.63} \\
			\midrule
			\midrule
			{\multirow{2}{*}{\specialcell{HABERTOR-\\VAFQF}}} 
			& {traditional}         
			& 92.89 & 86.42 & 79.64 
			& 97.42 & 88.08 & 79.71 \\
			\cline{2-8}
			& {\textbf{ours}}         
			& 93.51	& \textbf{87.38} & \textbf{ 80.16}
			& 97.37	& \textbf{88.21} & \textbf{80.23} \\
			\midrule
			\midrule
			{\multirow{2}{*}{\specialcell{HABERTOR-\\VAFOQF}}}
			& {traditional}         
			& 93.08 & 86.67 & 79.33 
			& 97.28 & 87.40 & 79.19 \\
			\cline{2-8}
			& {\textbf{ours}}         
			& 93.49 & \textbf{87.14} & \textbf{80.06}
			& 97.23	& \textbf{87.94} & \textbf{79.61} \\
			
			\bottomrule
		\end{tabular}
	}
	\vspace{-10pt}
\end{table}

\noindent{\bf Effectiveness of the adversarial attacking method FGM with our fined-grained and adaptive noise magnitude:} 
To show the effectiveness of the FGM attacking method with our proposed fine-grained and adaptive noise magnitude, we compare the performance of HABERTOR and its four factorized versions when (i) using a fixed and scalar noise magnitude as in the \textit{traditional} FGM method, and (ii) using a fine-grained and adaptive noise magnitude in \textit{our} proposal. We evaluate the results by performing the Language Model transfer learning on Twitter and Wiki datasets and present results in Table \ref{table:adv-comparison}. Note that, the noise magnitude range is set in [1, 5] for both two cases (i) and (ii) for a fair comparison, and we manually search the optimal value of the noise magnitude in the \textit{traditional} FGM method using the development set in each dataset. 
We observe that in all our five models, learning with our modified FGM produces better results compared to learning with a traditional FGM, confirming the effectiveness of our proposed fine-grained and adaptive noise magnitude.

\begin{figure}[t]
	\centering
	\begin{subfigure}{0.23\textwidth}
		\centering
		\includegraphics[width=\textwidth]{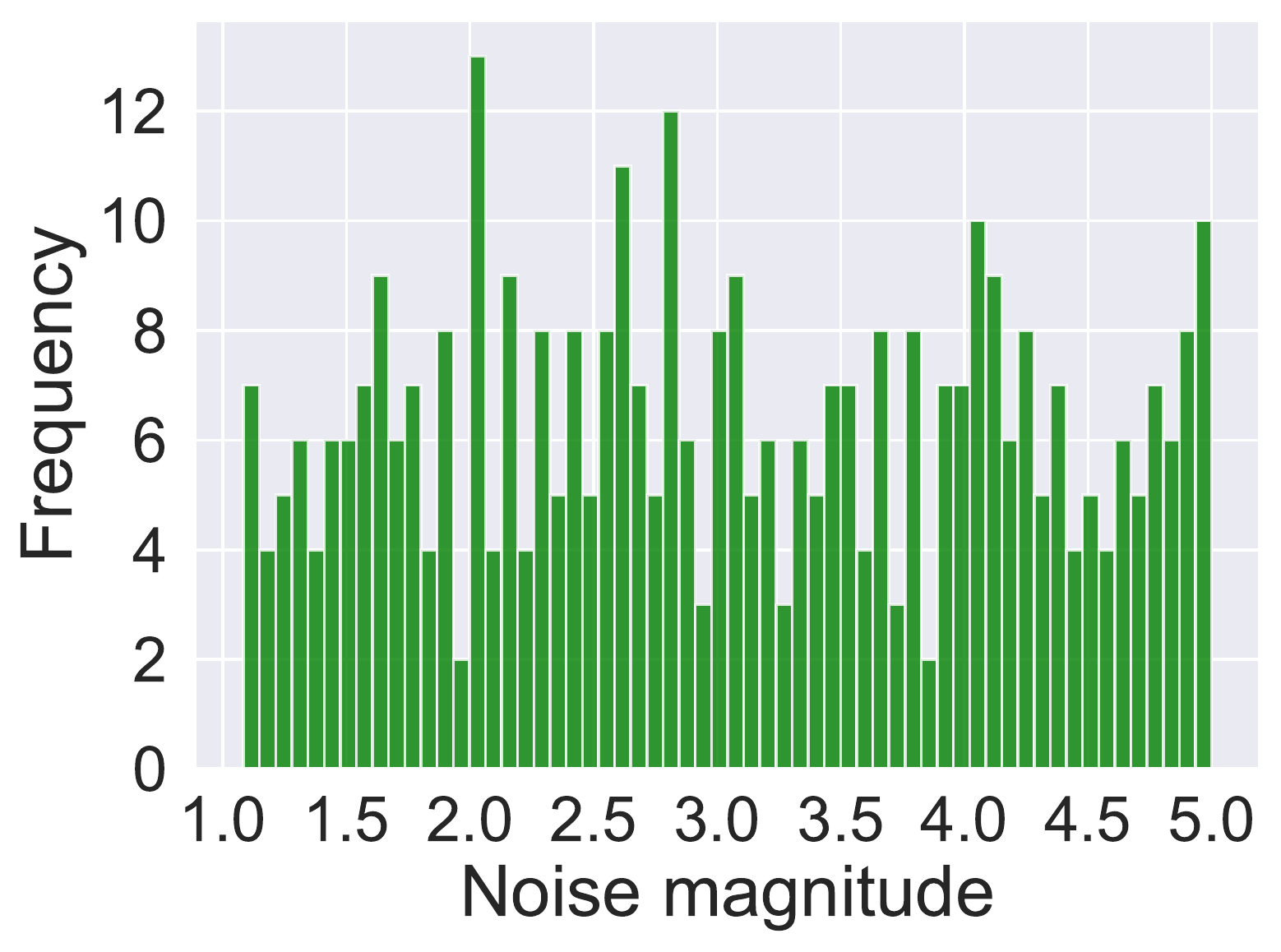}
		\caption{Twitter}
		\label{fig:noise-vis-twitter}
	\end{subfigure}
	\hfill
	\begin{subfigure}{0.23\textwidth}
		\centering
		\includegraphics[width=\textwidth]{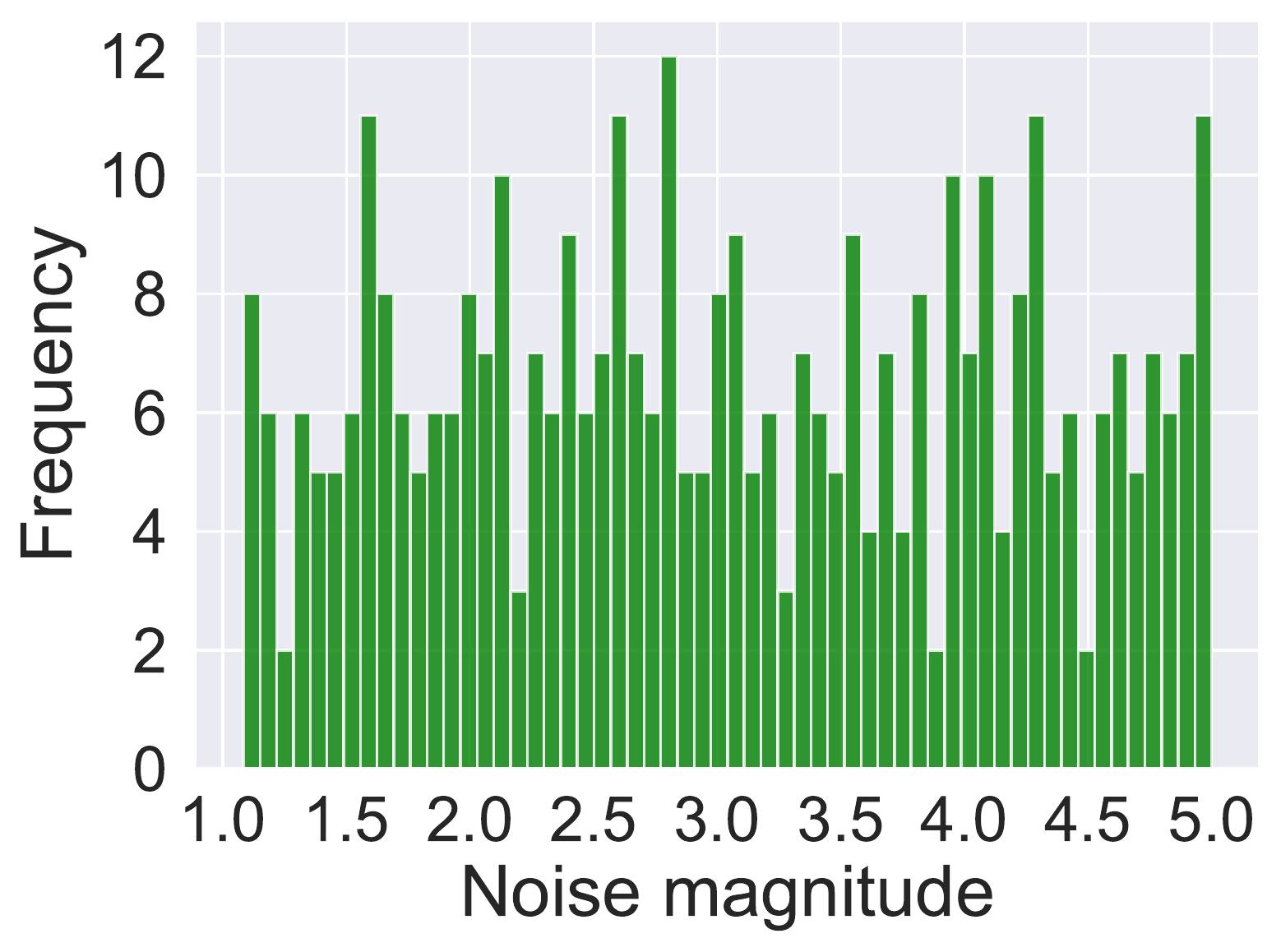}
		\caption{Wiki.}
		\label{fig:noise-vis-wiki}
	\end{subfigure}
	\vspace{-10pt}
	\caption{Histogram of the learned noise magnitude when performing Language Model transfer learning of HABERTOR on (a) Twitter, and (b) Wiki datasets.}
	\label{fig:noise-vis}
	\vspace{-15pt}
\end{figure}

We also plot the histogram of the learned noise magnitude of HABERTOR on Twitter and Wiki datasets. Figure \ref{fig:noise-vis} shows that different embedding dimensions are assigned with different learned noise magnitude, showing the need of our proposed fine-grained and adaptive noise magnitude, that automatically assigns different noise scales for different embedding dimensions.


\noindent\textbf{Additional Ablation study:} We conduct several ablation studies to understand HABERTOR's sensitivity. Due to space limitation, we summarize the key findings as follows, and leave detailed information and additional study results in the Appendix: 
(i) A large \emph{masking factor} in {HABERTOR} is helpful to improve its performance; 
(ii) Pretraining with a larger hatespeech dataset or a more fine-grained pretraining can improve the hatespeech prediction performance; 
and (iii) Our fine-tuning architecture with multi-source and ensemble of classification heads helps improve the performance.

\vspace{-3pt}
\subsubsection{Further application discussion}
\begin{table}
	\centering
	\caption{Application of our models on the sentiment classification task using Amazon Prime Pantry reviews.}
	\vspace{-10pt}
	\label{table:application}
	\resizebox{0.42\textwidth}{!}{
		\begin{tabular}{l|ccc}
			\toprule
			Model  & AUC & AP & F1 \\
			\midrule

ALBERT-base						  & 98.77 & 99.77 & 97.95 \\ 
BERT-base 			  				& 99.16 &  99.84 & 98.42 \\
			\midrule 
HABERTOR			 			   & 99.10 & 99.83 & 98.39 \\
HABERTOR+VQF 				  & 99.09 & 99.83 & 98.27 \\ 
HABERTOR+VAQF 				 & 98.90 & 99.80 & 98.07 \\ 
HABERTOR+VAFQF		 		& 98.87 & 99.79 & 98.05 \\ 
HABERTOR+VAFOQF 		  & 98.61 & 99.75 & 97.78 \\
			\bottomrule
		\end{tabular}
	}
	\vspace{-15pt}
\end{table}
Our proposals were designed for the hatespeech detection task, but in an extent, they can be applied for other text classification tasks. To illustrate the point, we experiment our models (i.e. all our pretraining and fine-tuning designs) on a sentiment classification task. Particularly, we used 471k Amazon-Prime-Pantry reviews \cite{mcauley2015image}, which is selected due to its reasonable size for fast pretraining, fine-tuning and result attainment.
After some preprocessings (i.e. duplicated reviews removal, convert the reviews with rating scores $\ge$ 4 as positive, rating $\le$ 2 as negative, and no neutral class for easy illustration), we obtained 301k reviews and splited into 210k-training/30k-development/60k-testing with a ratio 70/10/20. 
Next, we pretrained our models on 210k training reviews which contain 5.06M of words.  
Then, we fine-tuned our models on the 210k training reviews, selected a classification threshold on the 30k development reviews, and report AUC, AP, and F1 on the 60k testing reviews. We compare the performance of our models with fine-tuned BERT-base and ALBERT-base -- two best baselines.
We observe that though pretraining on only 5.06M words of 210k training reviews, HABERTOR performs very similarly to BERT-base, while improving over ALBERT-base. Except HABERTOR-VAFOQF with a little bit smaller F1-score compared to ALBERT-base, our other three compressed models worked better than ALBERT-base, showing the effectiveness of our proposals.

%% file: 7-Conclusion.tex
\vspace{-5pt}
\section{Conclusion}
\vspace{-3pt}
In this paper, we presented the \emph{HABERTOR} model for detecting hatespeech. \emph{HABERTOR} understands the language of the hatespeech datasets better, is 4-5 times faster, uses less than 1/3 of the memory, and has a better performance in hatespeech classification. 
Overall, \emph{HABERTOR} outperforms 15 state-of-the-art hatespeech classifiers and generalizes well to unseen hatespeech datasets, verifying not only its efficiency but also its effectiveness.



\vspace{-5pt}
\section*{Acknowledgments}
\vspace{-3pt}
This work was supported in part by NSF grant CNS-1755536.

%% file: 8-supplemental.tex
\section{Appendix}

\begin{figure*}[t]
	\centering
	\includegraphics[width=\textwidth]{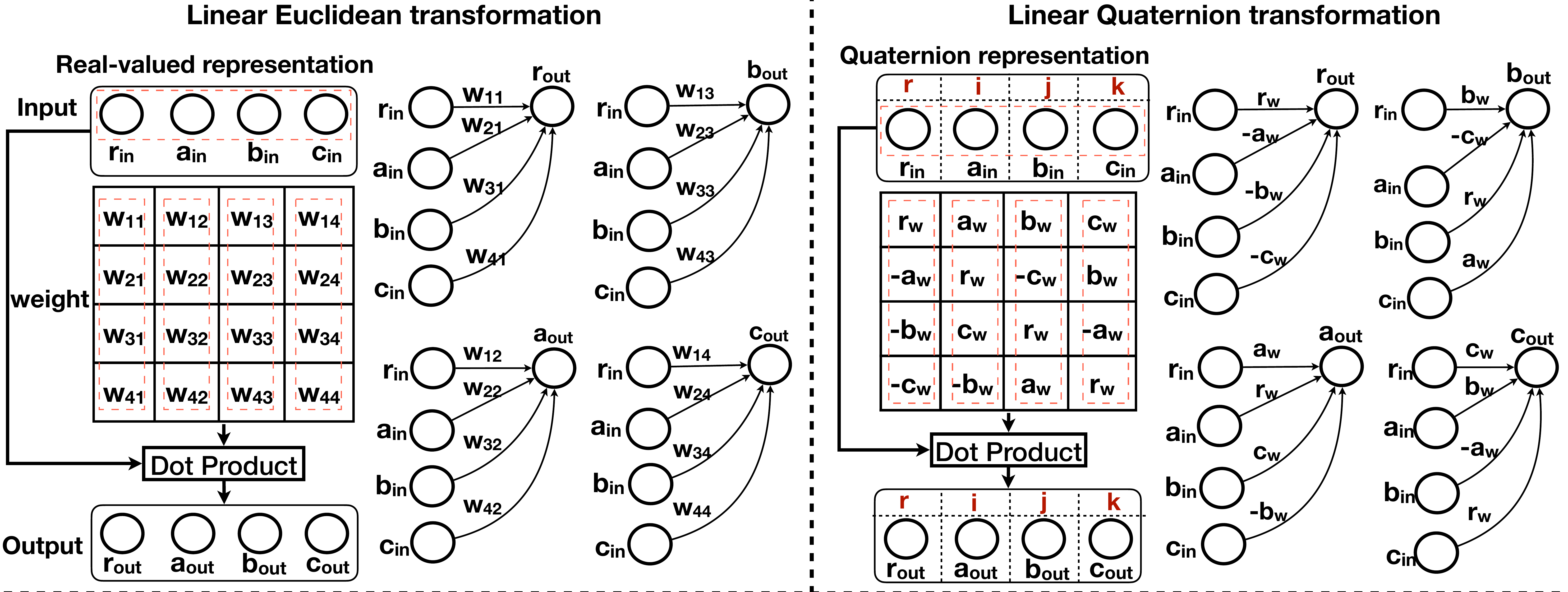}
	\vspace{-10pt}
	\caption{Comparison between linear Euclidean transformation (Left) and linear Quaternion transformation (Right). The Hamilton product in Quaternion space is replaced with an equivalent dot product in real space for an easy reference. Computing each output dimension in real-valued transformation (left) always need 4 new parameters, resulting in 16 degrees of freedom. In contrast, only 4 parameters are used and shared in producing all 4 output dimensions in Quaternion transformation, leading to a better inter-dependency encoding and a 75\% of parameter saving.
	}
	\vspace{-10pt}
	\label{fig:quaternions}
\end{figure*}

\begin{figure*}[t]
	\centering
	\includegraphics[width=\linewidth]{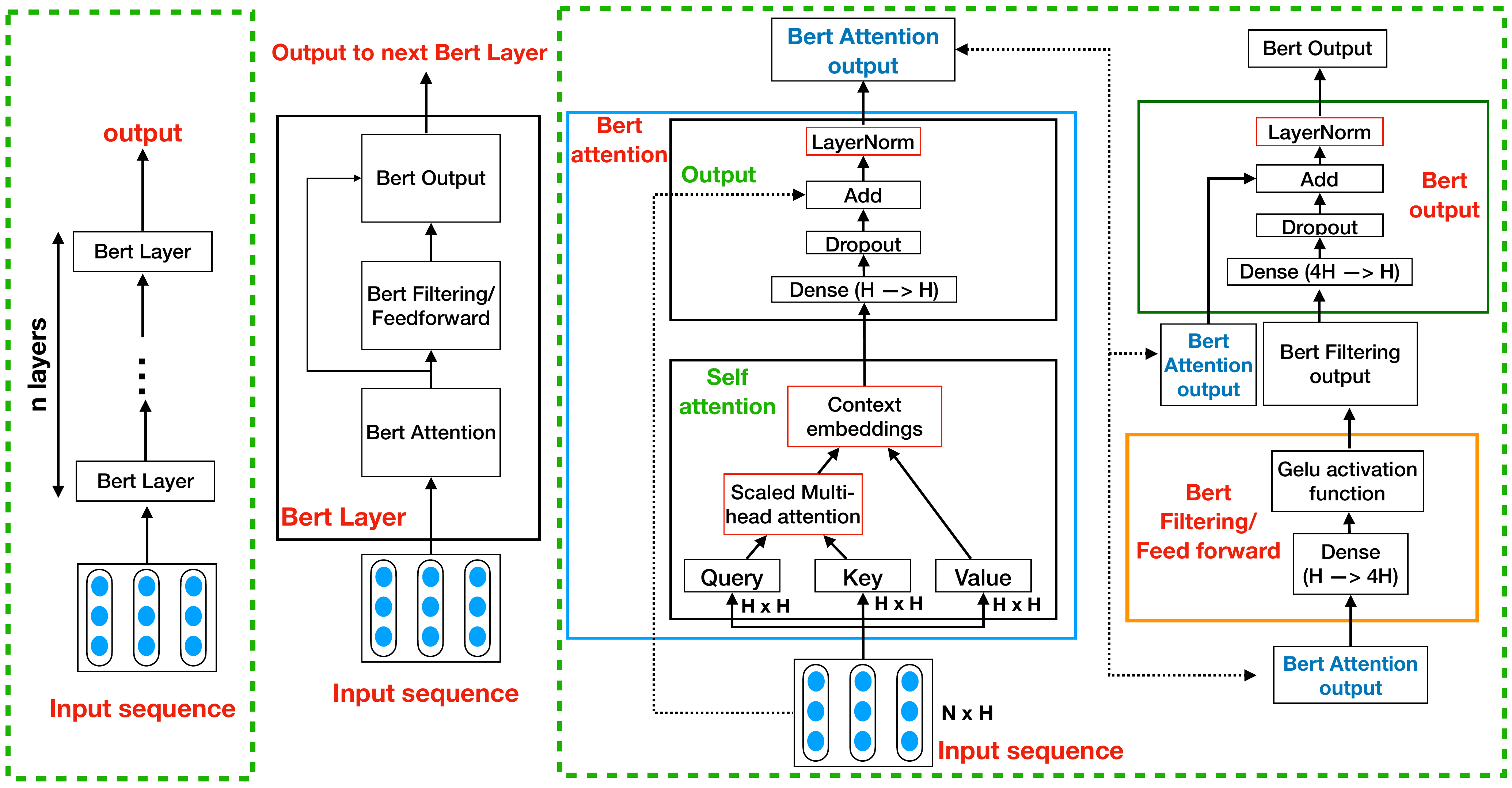}
	\caption{General view of the BERT architecture. Uncovering the architecture from left to right.}
	\label{fig:bert-architecture}
\end{figure*}

\subsection{Parameter Estimation for pretraining HABERTOR with language model tasks}
Given the following input sentences $s_i=[w_1^{(i)}, w_2^{(i)}, ..., w_u^{(i)}]$ and  $s_j=[w_1^{(j)},$ $w_2^{(j)},$ $..., w_v^{(j)}]$, let the text sequence be $c_l = s_{ij} = [w_1^{(i)}, w_2^{(i)}, ..., w_u^{(i)}, w_1^{(j)}, w_2^{(j)}, ..., w_v^{(j)}]$=$[w_1, ..., w_{n}]$ ($n=u+v$) with label $y_l$ where we already paired the sentences to generate a \emph{next} (i.e $y_l = 1$) or \emph{not next} (i.e. $y_l = 0$) training instance. Let $\bar{c}_l$ be a corrupted sequence of $c_l$, where we masked some tokens in $c_l$. Denote $\boldsymbol{C}$ a collection of such training text sequences $c_{l}$. The masked token prediction task aims to reconstruct each $c_l \in \boldsymbol{C}$ given the corrupted sequence $\bar{c_l}$. In another word, the masked token prediction task maximizes the following log-likelihood:
\vspace{-5pt}
\begin{equation*}
\resizebox{0.3\textwidth}{!}{$
\begin{aligned}
    \mathcal{L}_1 &= 
    \operatorname*{argmax}_{\theta} \; \sum_{l=1}^{|\boldsymbol{C}|} log \; p_\theta (c_l | \bar{c_l})  \\
    &\approx
    \sum_{l=1}^{|\boldsymbol{C}|} \sum_{t=1}^{n} \mathbbm{1}_t \; log \; p_\theta(w_t | \bar{c_l})
\end{aligned}
$}
\vspace{-5pt}
\end{equation*}
where $\mathbbm{1}_t$ is an indicator function and $\mathbbm{1}_t$ = 1 when the token $t^{th}$ is a \emph{[MASK]} token, otherwise $\mathbbm{1}_t$ = 0. $\theta$ refers to all the model's learning parameters, $w_t$ is the ground truth token at position $t^{th}$. Denote $H_\theta (c_l) = [H_\theta (c_l)_1, H_\theta (c_l)_2, ..., H_\theta (c_l)_n]$ as the sequence of transformed output embedding vectors obtaining at the final layer of corresponding $n$ tokens in the sequence $c_l$. $H_\theta (c_l)_t \in \mathcal{R}^d$ with $d$ is the embedding size. By parameterizing a linear layer with a transformation $W_1 \in \mathcal{R}^{V \times d}$ (with $V$ refers to the vocabulary size) as a decoder, we can rewrite $\mathcal{L}_1$ 
as follows:
\begin{equation*}
\resizebox{0.5\textwidth}{!}{$
\begin{aligned}
    &\mathcal{L}_1
    = \operatorname*{argmin}_{\theta}
    -\sum_{i=1}^{|\boldsymbol{C}|}
    \sum_{t=1}^{n}
    \mathbbm{1}_t \log\frac{\exp \bigg( \big\lbrack W_1H_\theta (\bar{c}_l)_t \big\rbrack _t \bigg)}
         {\sum_{k=1}^{V}
        \ exp \bigg( \big\lbrack W_1H_\theta (\bar{c}_l)_t \big\rbrack _k \bigg)}
\end{aligned}
$}
\end{equation*}
where $\lbrack \cdot \rbrack _t$ refers to the output value at position $t$.

For the next sentence prediction task, the objective is to minimize the following binary cross entropy loss function:
\vspace{-0.2cm}
\begin{equation*}
\resizebox{0.44\textwidth}{!}{$
\begin{aligned}
\mathcal{L}_2 = \operatorname*{argmin}_{\theta} \;  -\sum_{i=1}^{|\boldsymbol{C}|}
& y_l\ \log \big( \sigma (W_2 H_\theta (c_l)_1) \big) +  \\
& (1 - y_l) \log \big( \sigma (W_2 H_\theta (c_l)_1) \big)
\end{aligned}
$}
\vspace{-5pt}
\end{equation*}
\noindent where $W_2 \in \mathcal{R}^{d}$ and $H_\theta (c_l)_1$ refers to the embedding vector of the first token in the sequence $c_l$, or the \emph{[CLS]} token. The intuition behind this is that the \emph{[CLS]}'s embedding vector summarizes information of all other tokens via the attention Transformer network~\cite{vaswani2017attention}.

Then, pretraining with two language modeling tasks aims to minimize both loss functions 
$\mathcal{L}_1$ and $\mathcal{L}_2$
by:
$\mathcal{L}_{LM} = \operatorname*{argmin}_{\theta} \;  \big( \mathcal{L}_1 + \mathcal{L}_2 \big)$

\subsection{Quaternion}
In mathematics, Quaternions\footnote{https://en.wikipedia.org/wiki/Quaternion} are a hypercomplex number system. A Quaternion number \emph{P} in a Quaternion space $\mathbb{H}$ is formed by a real component (\emph{r}) and three imaginary components as follows:
\vspace{-6pt}
\begin{equation}
\label{equa:quaternion}
P = r + a \boldsymbol{i} + b \boldsymbol{j} + c \boldsymbol{k},
\end{equation}
where $\boldsymbol{ijk} = \boldsymbol{i^2} = \boldsymbol{j^2} = \boldsymbol{k^2} = -1$. The non-commutative multiplication rules of quaternion numbers are: $\boldsymbol{ij}=\boldsymbol{k}$, $\boldsymbol{jk}=\boldsymbol{i}$, $\boldsymbol{ki}=\boldsymbol{j}$, $\boldsymbol{ji}=-\boldsymbol{k}$, $\boldsymbol{kj}=-\boldsymbol{i}$, $\boldsymbol{ik}=-\boldsymbol{j}$. In Equa~(\ref{equa:quaternion}), $r, a, b, c$ are real numbers $\in \mathbb{R}$. Note that $r, a, b, c$ can also be extended to a real-valued vector $\in \mathbb{R}$ to obtain a Quaternion embedding, which we use to represent each word-piece embedding.

\noindent\textbf{Algebra on Quaternions:}
We present the Hamilton product on Quaternions, which is the heart of the linear Quaternion-based transformation. 
The Hamilton product (denoted by the $\otimes$ symbol) of two Quaternions $P \in \mathbb{H}$ and $Q \in \mathbb{H}$ is defined as:
\vspace{-4pt}
\begin{equation}
\label{equa:quaternion-hamiltonprod}
\begin{aligned}
P \otimes Q =& (r_P r_Q - a_P a_Q - b_P b_Q - c_P c_Q) + \\
             & (r_P a_Q + a_P r_Q + b_P c_Q - c_P b_Q) \boldsymbol{i} + \\
             & (r_P b_Q - a_P c_Q + b_P r_Q + c_P a_Q)  \boldsymbol{j} + \\
             & (r_P c_Q + a_P b_Q - b_P a_Q + c_P r_Q)  \boldsymbol{k} + \\
\end{aligned}
\end{equation}

\noindent\textbf{Activation function on Quaternions:} Similar to \cite{tay-etal-2019-lightweight,parcollet2018quaternion}, we use a \emph{split activation function} because of its stability and simplicity. \emph{Split activation function} $\beta$ on a Quaternion $P$ is defined as:
\vspace{-4pt}
\begin{equation}
\label{equa:quaternion-activation}
\beta(P) = f(r)  + f(a) \boldsymbol{i} + f(b)\boldsymbol{j} + f(c)\boldsymbol{k}
\end{equation}
, where $f$ is any standard activation function for Euclidean-based values.

\noindent\textbf{Why does a linear Quaternion transformation reduce 75\% of parameters compared to the linear Euclidean transformation?} 
Figure \ref{fig:quaternions} shows a comparison between a traditional linear Euclidean transformation and a linear Quaternion-based transformation.

In Euclidean space, the same input will be multiplied with different weights to produce different output dimensions. Particularly, given a real-valued 4-dimensional vector $[r_{in}, a_{in}, b_{in}, c_{in}]$, we need to parameterize a weight matrix of 16 parameters (i.e. 16 degrees of freedom) to transform the 4-dimensional input vector into a 4-dimensional output vector $[r_{out}, a_{out}, b_{out},$ $c_{out}]$. However, with Quaternion transformation, the input vector now is represented with 4 components, where $r_{in}$ is the value of the real component, $a_{in}$, $b_{in}$, $c_{in}$ are the corresponding value of the three imaginary parts $\boldsymbol{i}$, $\boldsymbol{j}$, $\boldsymbol{k}$, respectively. Because of the weight sharing nature of Hamilton product, different output dimensions take different combinations with the same input with exactly same 4 weighting parameters \{$r_{w}, a_{w}, b_{w}, c_{w}$\}. Thus, the Quaternions transformation reduces 75\% of the number of parameters compared to the real-valued representations in Euclidean space.

\noindent\textbf{Quaternion-Euclidean conversion:}
Another excellent property of using Quaternion representations and Quaternion transformations is that converting from Quaternion to Euclidean and vice versa are convenient. To convert a real-valued based vector $v \in \mathcal{R}^d$ into a Quaternion-based vector, we consider the first $d/4$ dimensions of $v$ as the value of the real component, and the corresponding next three $d/4$ dimensions are for the three imaginary parts, respectively. Similarly, to convert a Quaternion vector $v \in \mathcal{H}^d$ into a real-valued vector, we simply concatenate all four components of the Quaternion vector, and treat the concatenated vector as a real-valued vector in Euclidean space.

\subsection{Analysis on the BERT's Parameters}
\label{sec:bert-param-analysis}

Figure \ref{fig:bert-architecture} presents a general view of the BERT architecture. Each BERT layer contains three parts: (i) attention, (ii) filtering, and (iii) output. 

The \emph{attention} part parameterizes three weight transformation matrices H$\times$H to form key, query, and value from the input, and another weight matrix H$\times$H to transform the output attention results. The total parameters of this part are 4$H^2$. 
The \emph{filtering} part parameterizes a weight matrix H$\times$4H to transform the output of the \emph{attention} part, leading to a total of 4$H^2$ parameters. 
The \emph{output} part parameterizes a weight matrix 4H$\times$H to transform the output of the \emph{filtering} part from 4H back to H, resulting in 4$H^2$ parameters.

Thus, a BERT layer has 12$H^2$ parameters, and a BERT-base setting with 12 layers has 144$H^2$ parameters. By taking into account the number of parameters for encoding V vocabs, the total parameters of BERT is VH + 144$H^2$. 

\subsection{50-50 Rule}
To ensure the 50-50 rule, we perform the following method: Let M be the number of input text sequences that we can split into multiple sentences, and N be the number of input sequences that can be tokenized into only one sentence. We want the number of sentences to be generated as \emph{next} sentence pairs (sampling with probability $p_1$) to be roughly equal to the number of sentences to be formed as \emph{not next} sentence pairs (sampling with probability $p_2$). In another word, $M \times p_1 = (M + N)\times p_2$ or $\frac{p_1}{p_2} = \frac{(M + N)}{M}$. Since $p_1 + p_2 = 1$, replacing $p_2 = 1 - p_1$, we have: $M\times p_1 = (M + N) (1 - p_1) \longrightarrow p_1 = \frac{(M + N)}{(2M + N)}$. With $p_1$ established, we set $p_1$ as the probability for a sentence to be paired with another consecutive sentence in a same input sequence to generate a \emph{next} sentence example.

\subsection{Baselines and Hyper-parameter Settings}
15 Baselines are described as follows:
\squishlist
    \item {\bf BOW}: Similar to \citet{dinakar2011modeling,van2015automatic}, we extract bag of words features from input sequences, then a traditional machine learning classifier is built on top of the extracted features.

    \item {\bf NGRAM}: It is similar to BOW model, except using n-gram features of the input sequence.

    \item {\bf CNN} \cite{kim2014convolutional}: It is a state-of-the-art word based CNN neural network model.

    \item {\bf VDCNN} \cite{conneau2016very}: It is a character based CNN model with a deeper architecture and optional shortcut between layers.

    \item {\bf FastText} \cite{joulin2016fasttext}: An extension of the Word2Vec model, where it represents each word as an n-gram of characters to provide embeddings for rare words.

    \item {\bf LSTM} \cite{cho2014learning}: 
    We use: (i) the last LSTM output vector, and (ii) a pooling layer (\emph{max} and \emph{mean}) to aggregate LSTM output vectors and report only the best results.

    \item {\bf att-LSTM}: A LSTM model with an attention layer to aggregate LSTM hidden state vectors.

    \item {\bf RCNN} \cite{lai2015recurrent}: A combination between a bi-directional recurrent structure to capture contextual information and a \emph{max} pooling layer to extract key features.

    \item {\bf att-BiLSTM} \cite{lin2017structured}: It is a self-attentive Bidirectional LSTM model. 
    
    \item {\bf Fermi} \cite{indurthi-etal-2019-fermi}: The best hatespeech detection method, as reported in \cite{basile-etal-2019-semeval}. It built a SVM classifier on top of the pretrained embeddings from Universal Sentence Encoder (USE) \cite{cer2018universal} model.
    
    \item\textbf{Q-Transformer} \cite{tay-etal-2019-lightweight}: It is a Quaternion Transformer. It replaces all Euclidean embeddings and linear transformations by Quaternion emddings and Quaternion linear transformation. We use the \emph{full} version of Q-Transformer due to its high effectiveness.
    
    \item \textbf{Tiny-BERT} \cite{jiao2019tinybert}: It is a compressed model of BERT-base by performing knowledge distillation on BERT-base during its pretraining phase with smaller number of layers and embedding sizes. We adopt the Tiny-BERT 4 layers with 14.5M of parameters.
    
    \item \textbf{DistilBERT-base} \cite{sanh2019distilbert}: another knowledge distillation of the BERT-base model during the BERT's pre-training phase.
    
    \item \textbf{ALBERT-base} \cite{Lan2020ALBERT}: a light-weight version of BERT-base model with parameters sharing strategies and an inter-sentence coherence pretraining task.
    
    
    \item \textbf{BERT-base} \cite{devlin2018bert}: Similar to \citet{nikolov2019nikolov}, we use pre-trained BERT with 12 layers and uncased (our experiments show uncased works better than cased vocab) to perform fine-tuning for the hatespeech detection.

\squishend

For baselines that require word embeddings, to maximize their performances, we initialize word embeddings with both GloVe pre-trained word embeddings~\cite{Pennington14glove:global} and random initialization and report their best results. We implement BOW and NGRAM with Naive Bayes, Random Forest, Logistic Regression, and Xgboost classifiers, and then report the best results.

By default, our vocab size is set to 40k. The number of pretraining epochs is set to 60, and the batch size is set to 768.
The learning rate is set to 5e-5 for the masked token prediction and next sentence prediction tasks, which are the two pretraining tasks, and 2e-5 for the hatespeech prediction task, which is the fine-tuning task. The default design of {HABERTOR} is given at Figure \ref{fig:bert-flex-ensemble}, with one separated classification net with an ensemble of 2 heads for each input source. 
The \emph{masking factor $\tau$} is set to 10. 
The noise magnitude's bound constraint $[a, b]=[1,2]$ in Yahoo dataset, and $[a, b]=[1,5]$ in Twitter and Wiki datasets. $\lambda_{adv}$=1.0, and $\lambda_\epsilon$=1 in all three datasets.
We use \emph{min} pooling function to put a more stringent requirement on classifying hatespeech comments, as the number of hatespeech-labeled comments is the minority. All the pre-trained language models are fined-tuned with the Yahoo train set. For all other baselines, we vary the hidden size from \{96, 192, 384\} and report their best results.
We build VDCNN with 4 convolutional blocks, which have 64, 128, 256 and 512 filters with a kernel size of 3, and 1 convolution layer. Each convolutional block includes two convolution layers.
For {FastText}, we find that 1,2,3-grams and 1,2,3,4,5-character grams give the best performance. All models are optimized using Adam optimizer \cite{kingma2014adam}.

\subsection{Ablation Study}
\label{sec:ablation}
\begin{table*}[!h]
\centering
\caption{Ablation study of HABERTOR on Yahoo dataset (i.e. both Yahoo News + Finance, to save space). Default results are in bold. Better results compared to the default one are underlined.}
\label{table:ablation}
\resizebox{0.7\textwidth}{!}{
    \begin{tabular}{c|l|ccccc}
    \toprule
    \multirow{3}{*}{{\bf Goal}}  & \multicolumn{1}{c}{\multirow{2}{*}{Model}} & \multicolumn{5}{|c}{\bf Yahoo}                   \\
    & \multicolumn{1}{c}{}                       & \multicolumn{1}{|c}{AUC} & AP & \multicolumn{1}{c}{\begin{tabular}[c]{@{}c@{}}FPR@\\ 5\%FNR\end{tabular}} & \begin{tabular}[c]{@{}c@{}}FNR@\\ 5\%FPR\end{tabular} & F1  \\
    \midrule
    & Default
        & {\bf 94.77} & {\bf 72.35} & {\bf 26.11} & {\bf  25.08} & {\bf 66.18} \\
    \midrule
    \multirow{2}{*}{\textbf{AS1}}
    & - adv
        & 94.60 & 71.19 & 26.97 & 25.78 & 65.02  \\
    & - adv + $\tau=1$
        & 94.32 & 70.27 & 28.08 & 26.69 & 64.78   \\

    \midrule




    \multirow{2}{*}{{\bf AS2}}
    & - adv + $\tau=1$ under 250k data
            & 92.61 & 64.71 & 36.43 & 32.51 & 60.13 \\
    & - adv + $\tau=1$ under 500k data
            & 94.04 & 69.11 & 29.82 & 27.99 & 63.34 \\
     \midrule

    \multirow{3}{*}{{\bf AS3}}
    & + single source + single head
        & 94.70 & 71.82 & 26.82 & 25.55 & 65.16\\
    & + single head
        & 94.70 & 72.15 & 26.66 & 25.20 & 65.59 \\

    & + ensemble 4
        & \underline{94.78} & 72.18 & 26.29 & \underline{24.97} & 65.78 \\
    & + ensemble 8
        & 94.71 & 72.06 & 26.13 & 25.08 & 65.56 \\

     \midrule
    \multirow{1}{*}{{\bf AS4}}
    & - adv + $\tau=1$ - pretraining
            & 92.48 & 65.26 & 36.47 & 32.10 & 60.66\\
    \midrule

    \multirow{5}{*}{{\bf AS5}}
    & + 3 layers
        & 94.54 & 71.25 & 27.15 & 25.90 & 64.98\\
    & + 4 layers
        & 94.67 & 71.53 & 26.25 & 25.50 & 65.38\\
    & + 192 hidden size
        & 94.57 & 71.00 & 26.56 & 25.93 & 65.05\\
    & + 3 att heads
        & 94.69 & 72.00 & 26.72 & 25.43 & 65.75\\
    & + 4 att heads
        & 94.69 & 72.06 & 26.61 & 25.22 & 65.80\\
    & + 12 att heads
        & 94.70 & 72.01 & 26.28 & 25.14 & 65.64\\

    \midrule

    \end{tabular}
}
\end{table*}

\noindent{\bf Effectiveness of regularized adversarial training and masking factor $\tau$ (AS1):} Recall that by default, {HABERTOR} has 2 classification nets, each of the two nets has an ensemble of 2 classification heads, \emph{masking factor} $\tau=10$, and is trained with regularized adversarial training. {HABERTOR - \emph{adv}} indicates {HABERTOR} without regularized adversarial training, and {HABERTOR - \emph{adv} + $\tau$=1} indicates {HABERTOR} without regularized adversarial training and \emph{masking factor} $\tau$ of 1 instead of 10. Comparing {HABERTOR} with {HABERTOR - \emph{adv}}, we see a drop of AP by 1.16\%, F1-score by 1.16\%, and the average error rate increases by 0.78\% (i.e. average of FPR@5\%FNR and FNR@5\%FPR). This shows the effectiveness of additional regularized adversarial training to make {HABERTOR} more robust. 
%
%
%
Furthermore, comparing {HABERTOR - \emph{adv}} (with default $\tau$=10) with {HABERTOR - \emph{adv} + $\tau=1$}, we observe a drop of AP by 0.92\%, F1-score by 0.24\%, and an increment of average error rate by 1.01\%. This shows the need of both regularized adversarial training with our proposed fine-grained and adaptive noise magnitude, and a large \emph{masking factor} in {HABERTOR}.
%

\noindent{\bf Is pretraining with a larger domain-specific dataset helpful? (AS2):} We answer the question by answering a reverse question: does pretraining with smaller data reduce performance? We pre-train {HABERTOR} with 250k Yahoo comments data (4 times smaller), and 500k Yahoo comments data (2 times smaller). Then, we compare the results of HABERTOR \emph{- adv} + $\tau=1$ with HABERTOR \emph{- adv} + $\tau=1$ under 250k data, and HABERTOR \emph{- adv} + $\tau=1$ under 500k data. Table \ref{table:ablation} shows the results. We observe that pretraining with a larger data size increases the hatespeech prediction performance. We see a smaller drop when pretraining with 1M data vs 500k data (AP drops 0.6\%), and a bigger drop when pretraining with 500k data vs 250k data (AP drops 4.4\%).
We reason that when the pretraining data size is too small, important linguistic patterns that may exist in the test set are not fully observed in the training set. In short, pretraining with larger hatespeech data can improve the hatespeech prediction performance. 
Note that BERT-base is pre-trained on 3,300M words, which are 106 times larger than HABERTOR (only 31M words). Hence, the performance of HABERTOR can be boosted further when pre-training a hatespeech language model with a larger number of hateful representatives.

\noindent{\bf Usefulness of separated source prediction and ensemble heads (\textbf{AS3}):} We compare HABERTOR from \emph{Default} settings to using \emph{single source + single source} (i.e. one classification head for all data sources, see Figure~\ref{fig:bert-traditional}), \emph{single head} (i.e. multi-source and each source has a single classification head, see Figure~\ref{fig:bert-flex-ensemble}), and using more ensemble heads (i.e. multi-source + more ensemble classification heads, see Figure~\ref{fig:bert-flex-ensemble}).
Table \ref{table:ablation} shows that the overall performance order is \emph{multi-source + ensemble of 2 heads} $>$ \emph{multi-source + single head} $>$  \emph{single source + single head}, indicating the usefulness of our multi-source and ensemble of classification heads architecture in the fine-tuning phase. However, when the number of ensemble heads $\ge$ 4, we do not observe better performance.



\noindent{\bf Is pretraining two language modeling tasks helpful for the hatespeech detection task? (AS4)} We compare {HABERTOR-\emph{adv}} + $\tau=1$ with {HABERTOR-\emph{adv}} + $\tau=1$ - \emph{pretraining}, where we ignore the pretraining step and consider HABERTOR as an attentive network for pure supervised learning with random parameter initialization. In Table \ref{table:ablation}, the performance of HABERTOR without the language model pretraining is highly downgraded: AUC drops $\sim$-2\%, AP drops $\sim$-5\%, FPR and FNR errors are $\sim$+9\% and $\sim$+5\% higher, respectively, and F1 drops $~$-4\%. These results show a significant impact of the pretraining tasks for hatespeech detection.

\noindent{\bf Is HABERTOR sensitive when varying its number of layers, attention heads, and embedding size? (AS5)} In Table \ref{table:ablation}, we observe that {HABERTOR}+\emph{3 layers} and {HABERTOR}+\emph{4 layers} work worse than {HABERTOR} (6 layers), indicating that a deeper model does help to improve hatespeech detection. However, when we increase the number of attention heads from 6 to 12, or decrease the number of attention heads from 6 to 4, we observe that the performance becomes worse. We reason that when we set the number of attention heads to 12, since there is no mechanism to constrain different attention heads to attend on different information, they may end up focusing on the similar things, as shown in \cite{clark2019does}. But when reducing the number of attention heads to 4, the model is not complex enough to attend on more relevant information, leading to worse performance.
Similarly, when we reduce the embedding size from 384 in {HABERTOR} to 192, the performance is worse. Note that we could not perform experiments with larger embedding sizes and/or more number of layers due to high running time and memory consumption. However, we can see in Table~\ref{table:ablation} performance of smaller {HABERTOR} with 3 layers, 4 layers, or 192 hidden size still obtain slightly better than BERT-base results reported in Table \ref{table:yahoo-performance}. This again indicates the need for pretraining language models on hatespeech-related corpus for the hatespeech detection task.

\begin{figure}[t]
    \centering
    \begin{subfigure}{0.46\columnwidth}
        \centering
        \includegraphics[width=0.99\textwidth]{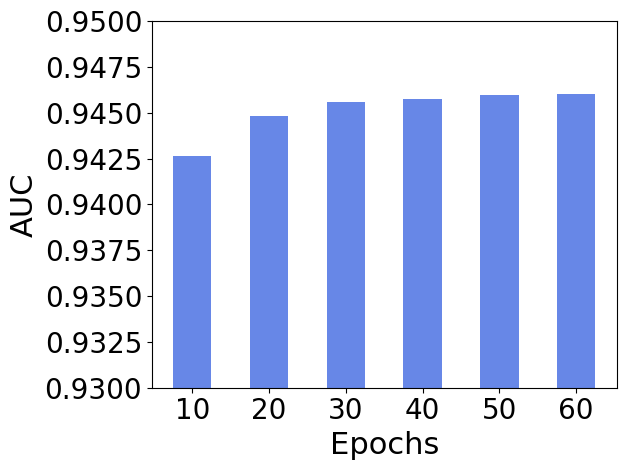}
        \vspace{-15pt}
        \caption{AUC.}
        \label{fig:AUC-pretraining}
    \end{subfigure}
    \hfill
    \begin{subfigure}{.46\columnwidth}
        \centering
        \includegraphics[width=0.99\textwidth]{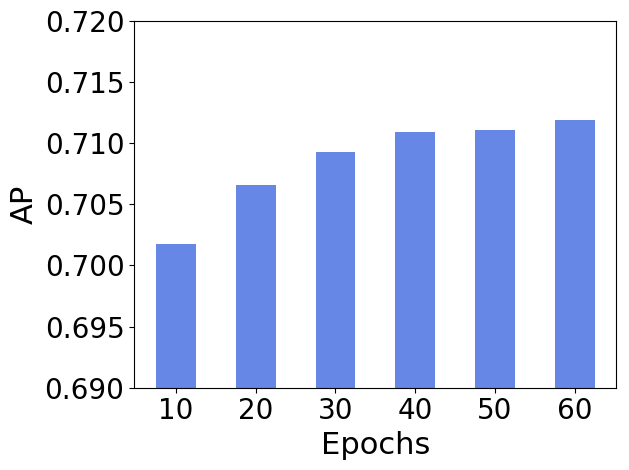}
        \vspace{-15pt}
        \caption{AP.}
        \label{fig:AP-pretraining}
    \end{subfigure}\hfill
    \vspace{-5pt}
    \caption{AUC and AP of HABERTOR without regularized adversarial training on Yahoo dataset when varying the number of epochs for the pretraining task.}
    \label{fig:compare-fine-grained}
    \vspace{-15pt}
\end{figure}
\noindent{\bf Effectiveness of fine-grained pretraining (AS6)?} Since the pretraining phase is unsupervised, a question is how much fine-grained pretraining should we perform to get a good hatespeech prediction performance? Or how many pretraining epochs are good enough? To answer the question, we vary the number of the pretraining epochs from \{10, 20, 30, ..., 60\} before performing the fine-tuning phase with hatespeech classification task. We report the changes in AUC and AP of fine-tuned {HABERTOR} on the Yahoo dataset without performing regularized adversarial training in Figure \ref{fig:compare-fine-grained}. We observe that a more fine-grained pretraining helps to increase the hatespeech prediction results, which is similar to a recent finding at \citet{liu2019roberta}, especially from 10 epochs to 40 epochs. However, after 40 epochs, the improvement is smaller.